%% file: main.tex
\documentclass[letterpaper]{article} 
\usepackage{arxiv}  
\usepackage{times}  
\usepackage{helvet}  
\usepackage{courier}  
\usepackage[hyphens]{url}  
\usepackage{graphicx} 
\usepackage{natbib}  
\usepackage{caption} 
\usepackage[linesnumbered,vlined,ruled]{algorithm2e}

\usepackage{booktabs}
\usepackage[normalem]{ulem}
\useunder{\uline}{\ul}{}
\usepackage{amsmath}
\usepackage{amssymb}
\usepackage{multirow}
\usepackage{bbding}
\usepackage[table,xcdraw]{xcolor}
\usepackage{fancyhdr}
\usepackage{fontawesome}

\usepackage{color} 
\usepackage{xcolor}
\usepackage{hyperref} 
\nocopyright 

\newcommand{\myparagraph}[1]{\noindent\textbf{#1}}
\newcommand{\methodname}{MagicMan}
\newcommand{\suppl}{\hyperref[sec:supp]{\textit{Supp. Mat.}}}
\newcommand{\thetitle}{MagicMan: Generative Novel View Synthesis of Humans \\ with 3D-Aware Diffusion and Iterative Refinement}

\definecolor{deeppink}{HTML}{FF1493}

\newcommand{\projectpage}
{\href{https://thuhcsi.github.io/MagicMan}{\textcolor{deeppink}{https://thuhcsi.github.io/MagicMan}}}

\title{
MagicMan: Generative Novel View Synthesis of Humans \\ with 3D-Aware Diffusion and Iterative Refinement
}
\author{
    Xu He\textsuperscript{\rm 1} \quad Xiaoyu Li\textsuperscript{\rm 2} \quad Di Kang\textsuperscript{\rm 2} \quad Jiangnan Ye\textsuperscript{\rm 1} \quad Chaopeng Zhang\textsuperscript{\rm 2} \\
    Liyang Chen\textsuperscript{\rm 1}  \quad Xiangjun Gao\textsuperscript{\rm 3} \quad Han Zhang\textsuperscript{\rm 4} \quad Zhiyong Wu\textsuperscript{\rm 1, 5, \Envelope} \quad Haolin Zhuang\textsuperscript{\rm 1}
}
\affiliations{
    \textsuperscript{\rm 1}Shenzhen International Graduate School, Tsinghua University \quad
    \textsuperscript{\rm 2}Tencent AI Lab \\
    \textsuperscript{\rm 3}The Hong Kong University of Science and Technology \quad
    \textsuperscript{\rm 4}Standford University \quad
    \textsuperscript{\rm 5}The Chinese University of Hong Kong \\
    \projectpage
}

\begin{document}
\thispagestyle{fancy}

\twocolumn[{%
\renewcommand\twocolumn[1][]{#1}%
\maketitle
\begin{center}
    \centering
    \captionsetup{type=figure}
    \includegraphics[width=\textwidth]{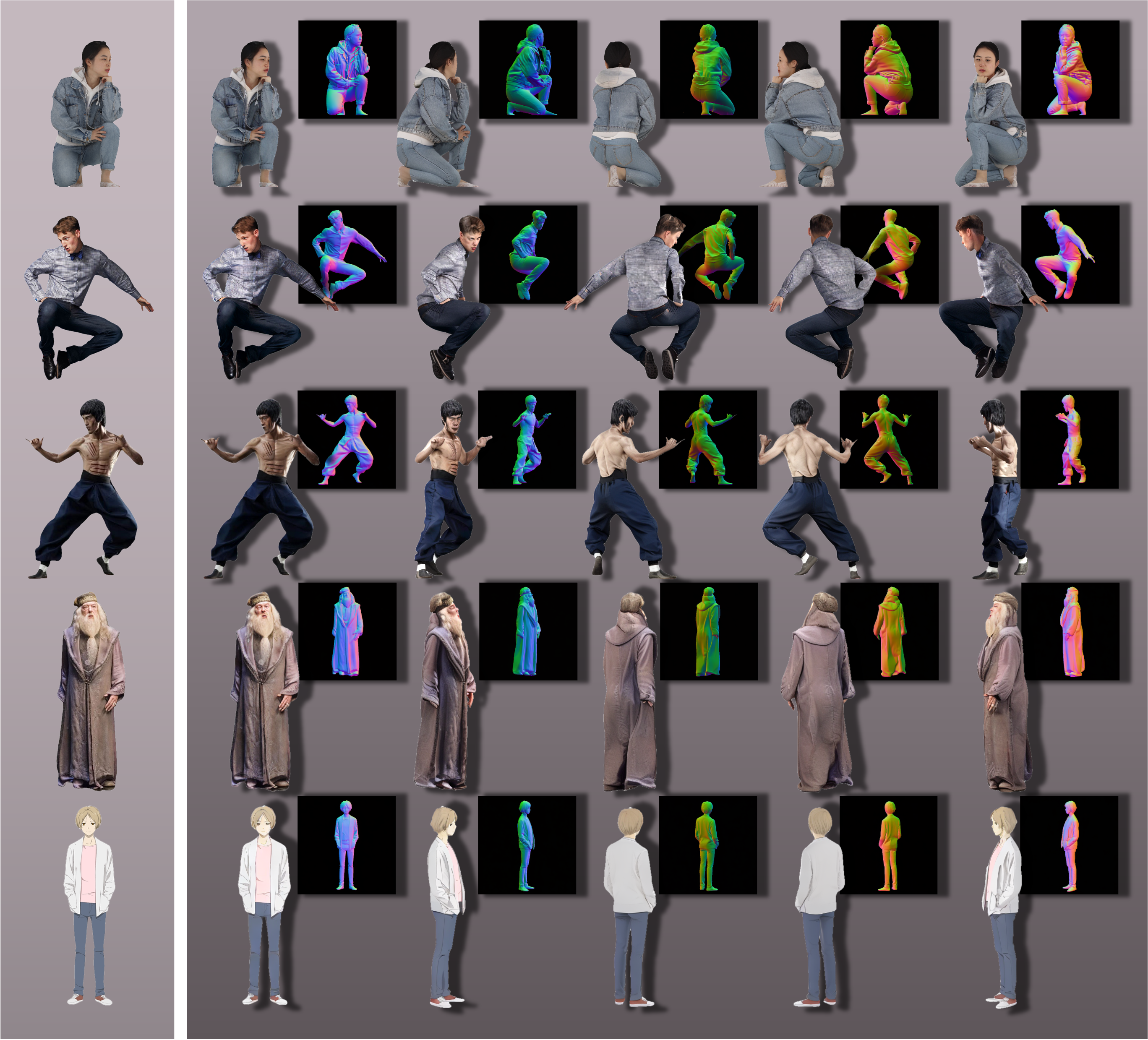}
    \caption{
    Given a single reference human image in different poses, outfits, or styles (i.e. real and fictional characters) as input, 
    \textbf{\methodname} is able to generate consistent high-quality novel view images and normal maps, which are well-suited for downstream multi-view reconstruction applications.
    }
    \label{fig:teaser}
\end{center}%
}]

\input{sec/body}

\bibliography{arxiv}

\input{sec/supp}

\end{document}

%% file: sec/body.tex
\begin{abstract}
Existing works in single-image human reconstruction suffer from weak generalizability due to insufficient training data or 3D inconsistencies for a lack of comprehensive multi-view knowledge. 
In this paper, we introduce \textbf{\methodname{}}, a human-specific multi-view diffusion model designed to generate high-quality novel view images from a single reference image. 
As its core, we leverage a pre-trained 2D diffusion model as the generative prior for generalizability, with the parametric SMPL-X model as the 3D body prior to promote 3D awareness. 
To tackle the critical challenge of maintaining consistency while achieving dense multi-view generation for improved 3D human reconstruction, we first introduce hybrid multi-view attention to facilitate both efficient and thorough information interchange across different views.
Additionally, we present a geometry-aware dual branch to perform concurrent generation in both RGB and normal domains, further enhancing consistency via geometry cues. 
Last but not least, to address \emph{ill-shaped} issues arising from inaccurate SMPL-X estimation that conflicts with the reference image, we propose a novel iterative refinement strategy, which progressively optimizes SMPL-X accuracy while enhancing the quality and consistency of the generated multi-views. 
Extensive experimental results demonstrate that our method significantly outperforms existing approaches in both novel view synthesis and subsequent 3D human reconstruction tasks.
\end{abstract}

\section{Introduction}
3D digital human creation is an important technique in computer vision and graphics, with a wide range of applications in games, movies, virtual and augmented reality. 
Traditional methods~\cite{balan2007detailed, bhatnagar2019multi, vlasic2009dynamic} usually utilize a dense camera array to capture synchronized posed multi-view images for human reconstruction.
However, these methods typically require a tedious and time-consuming process which is not practical for general users. 
Therefore, creating 3D human models from a single reference image is a significant task to be investigated.

To this end, early works like  PIFu~\shortcite{saito2019pifu}, PaMIR~\shortcite{zheng2021pamir}, ICON~\shortcite{xiu2022icon}, and ECON~\shortcite{xiu2023econ} have been introduced to train feed-forward networks on scanned 3D human datasets, capable of reconstructing 3D humans from a single image. 
Nevertheless, data for 3D human scans are extremely scarce with limited diversity, and thus these methods exhibit poor generalizability to challenging poses and outfits. 
Besides, the weak generative capability from insufficient data also leads to overly-smoothed geometry and blurred textures.

Thanks to the abundant priors in text-to-image diffusion models trained on large-scale data, another category of works~\cite{zhang2024humanref,gao2024contex,huang2024tech} leverage pre-trained diffusion models to optimize 3D representations through a Score Distillation Sampling (SDS) loss to create human models from a single image.
Although these approaches yield impressive results in terms of generalization and detailed textures, the absence of 3D awareness during SDS frequently leads to 3D inconsistencies like multi-face Janus problem~\shortcite{poole2022dreamfusion}.
Moreover, text descriptions or CLIP embeddings of reference images, which are typically used to guide SDS, contain only global semantic information and thereby hinder the generation of novel views with fine-grained textures consistent with the reference image. 

In the field of 3D object generation, as exemplified by Zero123~\shortcite{liu2023zero}, efforts have been made to use image diffusion models trained on multi-view data to directly generate novel views from a single image, which are subsequently used to reconstruct 3D models,  achieving promising results.

\begin{figure}[t]
\centering
\includegraphics[width=\linewidth]{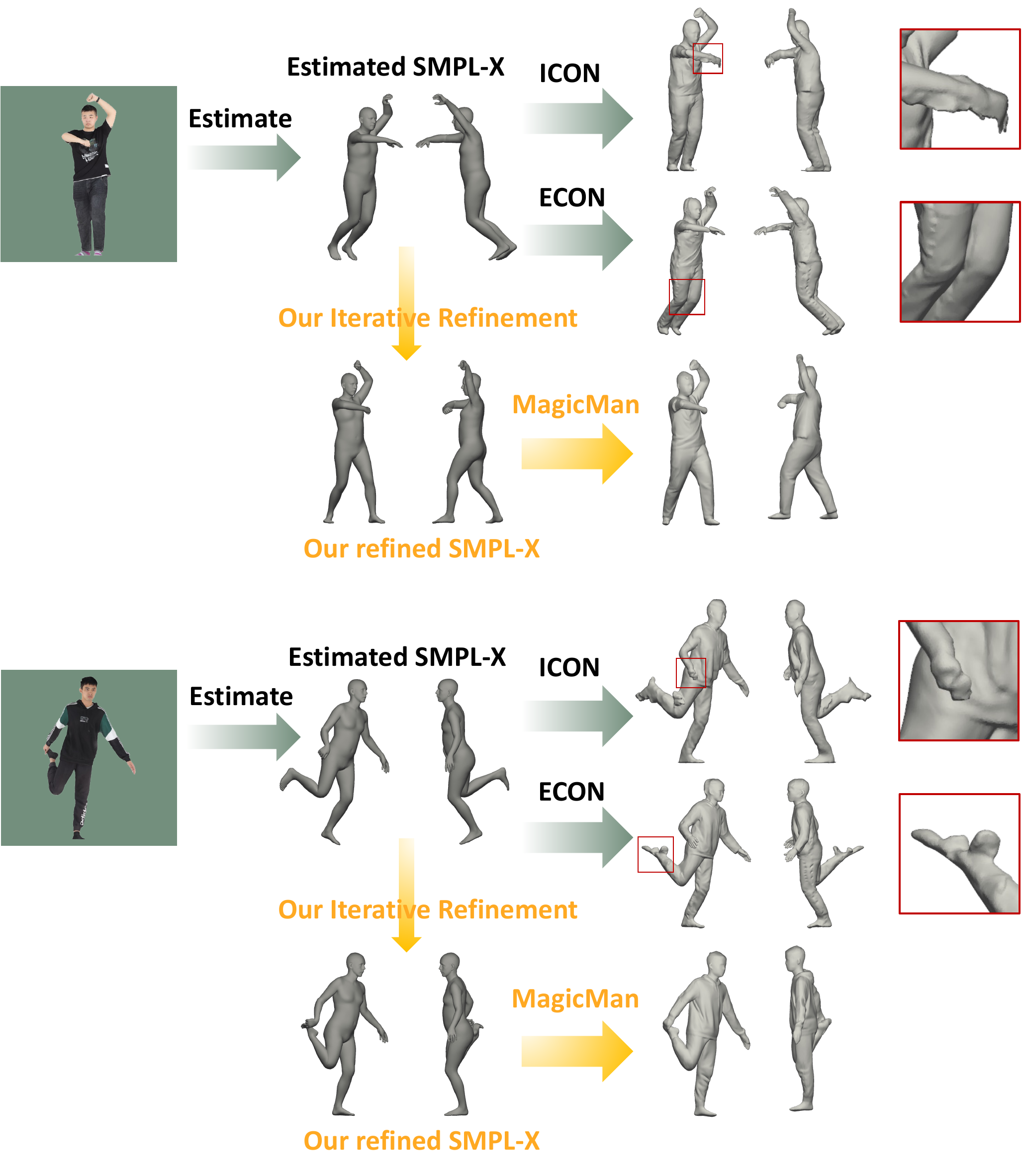}
\caption{\textbf{Typical \emph{ill-shaped} geometry caused by inaccurate SMPL-X poses} in recent 3D human reconstruction works ICON~\shortcite{xiu2022icon} and ECON~\shortcite{xiu2023econ}: 1) abnormal overall poses like tilted body (upper); 2) broken and distorted body parts (lower). Our proposed iterative refinement significantly improves the accuracy of SMPL-X poses, alleviating \emph{ill-shaped} problems.
}
\label{fig:ill}
\end{figure}

Following this paradigm, we present a multi-view diffusion model to directly generate human novel views, encompassing both generalizability and multi-view knowledge for 3D consistency.
There are three main challenges in this task:
Primarily, ensuring consistency for human multi-view images presents greater complexity than for 3D objects, due to the intricate geometry and detailed textures of human models.
Secondly, existing works enhance multi-view consistency either by extending 2D self-attention in image diffusion models to 3D attention across all views~\cite{shi2023mvdream, long2024wonder3d}, or by integrating 3D representations~\cite{liu2023syncdreamer}. Both approaches are memory-consuming and thus can only generate sparse consistent views, which are insufficient for reliable reconstruction of 3D human models where self-occlusion typically occurs. 
Finally, recent researches in 3D human reconstruction and generation~\cite{xiu2022icon,xiu2023econ,zheng2021pamir,liu2024humangaussian,huang2024tech} demonstrate the importance of utilizing parametric human models, e.g., SMPL~\shortcite{SMPL:2015} or SMPL-X~\shortcite{pavlakos2019expressive}, as 3D body priors, which is also confirmed to promote multi-view consistency in our task. However, SMPL-X meshes estimated from monocular images often display inaccurate poses, featuring depth ambiguities and misalignment with the reference image. This pose-image mismatch typically leads to \emph{ill-shaped} geometry, manifesting as abnormal overall poses or distorted body parts as illustrated in Fig.~\ref{fig:ill}.

To address these challenges, we propose \textbf{\methodname}, which efficiently produces high-quality, dense, and consistent multi-view images for humans from a single reference image. 
Our core insight lies in utilizing Stable Diffusion (SD)~\cite{rombach2022high} as the 2D prior for generation and SMPL-X mesh as the 3D body prior for better consistency. 
Therefore, we present a view-conditioned diffusion model with the reference image and SMPL-X guidance as the network backbone. 
To obtain dense and consistent novel views, we first introduce a hybrid multi-view attention mechanism to establish connections between different views. 
Specifically, it includes an efficient
1D attention across all views, and a 3D attention that spans both spatial and view dimensions operating on a sparse subset of selected views to enhance information interchange with minimal memory overhead. This hybrid mechanism facilitates thorough information fusion between views with reduced memory consumption, enabling us to generate significantly denser (i.e., 20 views at 512$\times$512 resolution) novel views while maintaining consistency.
Next, to deal with detailed geometry, we propose a geometry-aware dual branch with shared blocks to simultaneously generate aligned novel view RGB images and normal maps. The normal map prediction supplements structure information, further improving the consistency of geometric details.
Last but not least, to address the \emph{ill-shaped} issues arising from inaccurate SMPL-X estimates, we propose an iterative refinement approach, wherein intermediate generated multi-views are employed to refine the SMPL-X parameters for more accurate poses, which in turn guide the multi-view generation process with more consistent results.

To summarize, our main contributions are as follows:
\begin{itemize}
    \item We present \textbf{\methodname}, a method designed to generate high-quality multi-view images for humans from a single reference image, thereby facilitating seamless 3D human reconstruction.
    \item We propose an efficient hybrid multi-view attention to generate denser multi-view human images while maintaining better 3D consistency.
    \item A geometry-aware dual branch is introduced to perform generation in both RGB and normal domains, further enhancing multi-view consistency via geometry cues. 
    \item An iterative refinement strategy is proposed to progressively enhance both the SMPL-X pose accuracy and the generated multi-view consistency, reducing \emph{ill-shaped} issues arising from unreliable SMPL-X estimation.
\end{itemize}

\section{Related Work}
\label{sec:relatedwork}
\begin{figure*}[t]
\centering
\includegraphics[width=\textwidth]{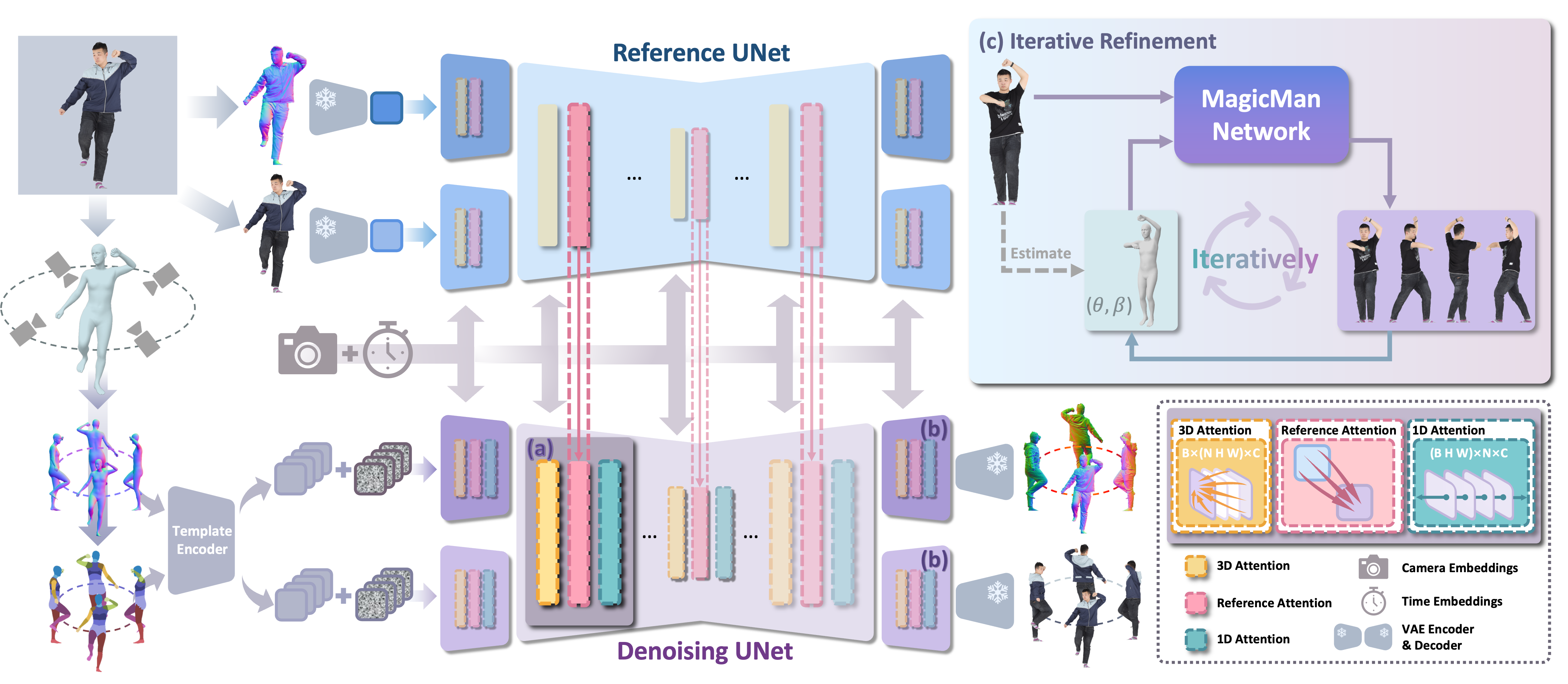}
\caption{Given a single human image, our proposed \textbf{\textbf{\methodname}} utilizes a pre-trained 2D diffusion model with a 3D human body prior to generate novel view images for humans. First, the reference image is fed into the denoising UNet via a reference UNet, with the viewpoint condition incorporated through camera embeddings. 
The rendered normal and segmentation maps of the posed SMPL-X mesh that corresponds to the reference image are also provided as geometry guidance to facilitate 3D awareness and consistency.
To obtain dense and consistent novel view images, we modify the attention module to a more efficient hybrid 1D-3D attention (a) to establish comprehensive connections between multi-views, and propose a geometry-aware dual branch (b) to also generate normal images complementary to RGB images via geometry cues. 
Last but not least, a novel iterative refinement strategy (c) is proposed in the inference stage to gradually update the initially estimated inaccurate SMPL-X pose and the synthesized novel view images, substantially reducing the \emph{ill-shaped} issues arising from unreliable SMPL-X estimates.
}
\label{fig:pipeline}
\end{figure*}

Our work focuses on the generative novel view synthesis from a single human image using diffusion models, which is related to the research topics in diffusion models, generative view synthesis, and human image synthesis. 

\noindent\textbf{Diffusion Models.}
Diffusion models~\shortcite{sohl2015deep, ho2020denoising} have demonstrated promising results in recent image synthesis works~\cite{dhariwal2021diffusion}. Classifier guidance~\shortcite{dhariwal2021diffusion} is proposed to boost sample quality with a trained classifier for conditional generation, while~\citeauthor{ho2022classifier} performs classifier-free guidance by combining the score estimates from conditional and unconditional generation. Based on diffusion models and large-scale data, various image synthesis tasks have achieved impressive results such as text-to-image synthesis~\cite{nichol2021glide, saharia2022photorealistic, ramesh2022hierarchical, rombach2022high} and super-resolution~\shortcite{saharia2022image}. In this paper, we explore using diffusion models for human novel view synthesis.

\noindent\textbf{Generative Novel View Synthesis.}
Generative novel view synthesis requires synthesizing views far beyond the input. Compared with traditional regression-based methods~\cite{yu2021pixelnerf, wang2021ibrnet}, it only has sparse or a single view as input to hallucinate unseen parts. With the development of generative models, this task has been studied by utilizing convolutional networks~\cite{zhou2016view}, generative
adversarial networks~\cite{li2022infinitenature, liu2021infinite, koh2021pathdreamer} and transformer~\cite{kulhanek2022viewformer, sajjadi2022scene}. Recently, diffusion models have also been applied to this task. Zero-1-to-3~\shortcite{liu2023zero} proposes a viewpoint-conditioned diffusion model trained on a large-scale dataset that shows a strong zero-shot generalizability. To incorporate 3D information into the 2D diffusion models, GeNVS~\shortcite{chan2023generative} and SyncDreamer~\shortcite{liu2023syncdreamer} use a 3D feature as  condition.~\citeauthor{tseng2023consistent} uses an epipolar attention for consistent view synthesis. Wonder3D~\shortcite{long2024wonder3d} extend diffusion models with multi-view and cross-domain attention and SV3D~\shortcite{voleti2024sv3d} utilizes a video diffusion model to improve consistency. In this work, we use diffusion models combined with the parametric body model SMPL-X~\shortcite{pavlakos2019expressive} as priors to ensure consistent view synthesis of humans. 

\noindent\textbf{Human Image Synthesis.}
Human image synthesis aims to synthesize novel views or poses giving a source image as inference. This problem is explored by using generative adversarial networks to synthesize the novel pose results in a single forward pass~\cite{ma2017pose, esser2018variational, siarohin2018deformable, li2019dense, zhu2019progressive, ren2020deep, men2020controllable, zhang2021pise, lv2021learning, zhou2022cross, ren2022neural} while Liquid Warping GAN~\shortcite{liu2019liquid} also demonstrates the human novel view synthesis results. 
However, GAN-based methods require transferring the style from the source to the target image in a single forward pass and are hard to capture complex structures~\shortcite{bhunia2023person}. On the other hand, diffusion models simplify the generation process into several denoising steps and therefore have been used for human image synthesis recently. \cite{bhunia2023person} introduces the first diffusion-based approach for pose-guided person synthesis. In addition, DreamPose~\shortcite{karras2023dreampose}, Animate Anyone~\shortcite{hu2024animate}, MagicAnimate~\shortcite{xu2024magicanimate}, and Champ~\shortcite{zhu2024champ} generates animated videos from a source image using pre-trained SD.
These methods could synthesize human images in novel views by giving the corresponding poses. However, without considering the 3D information between different views, it is hard to ensure view consistency. To address this problem, we utilize SMPL-X model to guide the synthesis process and explore more appropriate attention for multi-view consistency.

\section{Method}
Our proposed \textbf{\methodname{}}, as illustrated in Fig.~\ref{fig:pipeline}, takes a single reference human image as input and generates high-quality consistent dense (i.e., 20 views) multi-view images.
To utilize human priors from abundant in-the-wild images, \methodname{} adopts a pre-trained diffusion model~\shortcite{rombach2022high} as the backbone, and accepts one reference image along with SMPL-X pose and viewpoint as generation conditions (Sec.~\ref{subsec:method-guidance}).
We modify the original diffusion blocks with an efficient hybrid-attention mechanism connecting different views, which contains a 1D attention across all views and a 3D attention across spatial locations and sparse selected views (Sec.~\ref{subsec:method-hybrid-attention}).
To achieve better geometric consistency, we propose a geometry-aware dual branch complementary to novel view image synthesis (Sec.\ref{subsec:dual-branch}).
Last but not least, we propose a novel iterative refinement strategy by updating both the accuracy of SMPL-X poses and the quality of generated multi-views across iterations, reducing the \emph{ill-shaped} issues resulting from inaccurate pose estimation (Sec. \ref{subsec:method-optimization}). 

\subsection{Conditional Diffusion Model}
\label{subsec:method-guidance}
Our backbone is a denoising UNet~\shortcite{ronneberger2015u} that inherits the structure and pre-trained weights from SD 1.5~\shortcite{rombach2022high}. 
The vanilla SD UNet consists of downsampling, middle, and upsampling blocks.
Each block contains several interleaved convolution layers, self-attention layers that perform feature aggregation across spatial locations, and cross-attention layers that interact with CLIP text embeddings.
In our work, the denoising UNet is required to take multiple noise latents as input and produce human images of generated views consistent with the reference. 
Therefore, we incorporate the reference image and viewpoint information into the network, and provide SMPL-X meshes as geometry guidance to promote 3D awareness and consistency.

\myparagraph{Reference UNet.}
Inspired by recent advances in character animation~\cite{hu2024animate,xu2024magicanimate,zhu2024champ}, 
we utilize a copy of the denoising UNet to extract features from reference images, which we refer to as the reference UNet, ensuring consistency between the generated images and the reference image at both semantic and low levels. 
Different from previous works that integrate reference information into self-attention and retain CLIP embeddings for cross-attention, we find CLIP to be redundant and simply discard it. Instead, we replace cross-attention with reference attention, while leaving self-attention unaltered.

Let $x \in \mathbb{R}^{B \times N \times H \times W \times C}$ and $y \in \mathbb{R}^{B \times H \times W \times C}$ denote the corresponding feature maps from denosing UNet and reference UNet respectively, with batch size $B$, view count $N$, spatial size $H \times W$, and number of channels $C$.
Since the reference image is shared for all views, feature maps $y$ from the reference net are replicated $N$ times.
And then both $x$ and $y$ are reshaped to $\mathbb{R}^{(BN)\times (HW)\times C}$ for the following attention calculation.
Mathematically,
\begingroup
\fontsize{9}{12}\selectfont
\begin{align}
Q^{\text{ref}}=W^{\text{ref}}_Q x, K^{\text{ref}}=W^{\text{ref}}_K \left(x \oplus y\right), V^{\text{ref}}=W^{\text{ref}}_V \left(x \oplus y \right), \label{eq:ref-attn-qkv}
\end{align}
\endgroup
where $\oplus$ denotes concatenation along the $HW$ dimension.

\myparagraph{Pose guidance and viewpoint control.} 
A parametric SMPL-X mesh is estimated for the reference image and rendered according to the generated viewpoints to serve as pose and viewpoint conditions. 
We render normal and segmentation maps, which are encoded with a 4-layer convolution template encoder~\cite{hu2024animate} and added to the latent noise, to provide complementary geometric and semantic information~\cite{zhu2024champ}.
In addition, view control is explicitly incorporated into the network through camera embeddings via an MLP and added to the denoising time embeddings.

\subsection{Hybrid Multi-View Attention}
\label{subsec:method-hybrid-attention}
With the reference UNet, we can generate independent images consistent with the reference. 
Next, we want to establish connections across different views to produce consistent multi-views. 
To generate as many views as possible to capture comprehensive human information while maintaining consistency, the key problem lies in how to ensure thorough information exchange across a wide range of views in a memory-efficient manner.
Therefore, we propose a novel hybrid attention mechanism that combines the strengths of two types of multi-view attention, i.e. efficiency of 1D attention and thoroughness of 3D attention in multi-view interaction.

\myparagraph{1D multi-view attention}. 
First, we insert an additional 1D attention layer behind the reference attention to establish connections between different views. This module enhances the multi-view similarity in a highly memory-efficient manner, as it is calculated along view dimension only between identical pixel locations, allowing coherent generation of up to 20 views in a single forward pass.
Specifically, the feature map is reshaped to $\mathbb{R}^{(BHW)\times N \times C}$ to calculate self-attention along $N$, and we employ relative positional encoding~\shortcite{hwang2022tutel} instead of the commonly used sinusoidal encoding to account for relative viewpoint differences.

\myparagraph{3D multi-view attention.} 
Relying solely on 1D attention leads to content drift issues~\shortcite{shi2023mvdream} between views after large viewpoint changes (Fig.~\ref{fig:ablation-nvs}) since 1D attention lacks interaction between pixels at different locations and cannot find the corresponding pixel from other views.
Therefore, we further integrate 3D multi-view attention, facilitating more thorough information sharing across both spatial and view dimensions. 
Owing to the initial interactions established by 1D attention, 3D attention can be confined to a sparse subset of views without incurring excessive memory overhead.

To be specific, we extend the origin self-attention of denoising UNet to 3D attention. For a reshaped feature map $x_i \in \mathbb{R}^{B\times(HW)\times C}$ of each view, 3D attention is efficiently conducted with a small number of feature maps $x_{j_{1}:j_{M}} \in \mathbb{R}^{B\times (MHW) \times C}$ in $M$ views selected from the other views, calculated between all pixels across $x_i$ and $x_{j_{1}:j_{M}}$ with:
\begin{equation}
\fontsize{8}{12}
Q=W_Q x_i, 
K=W_K \left(x_i \oplus x_{j_{1}:j_{M}}\right), \\
V=W_V \left(x_i \oplus x_{j_{1}:j_{M}} \right).
\end{equation}

More complete connections between different views are established by our hybrid 1D-3D attention without overwhelming computational cost, 
enabling the generation of dense and consistent multi-views. 
In practice, the sparse subset of views selected for 3D attention vary across different UNet blocks, making full use of different levels of information, as detailed in \suppl

\newcommand{\mycommentstyle}[1]{\fontsize{8pt}{10pt}\ttfamily{#1}} 
\SetCommentSty{mycommentstyle}
\begin{algorithm}[tb]
\footnotesize
\caption{Iterative Refinement}
\label{alg:iterative-template-optimization-feedback}
\SetKwInOut{Parameter}{Parameter}
\KwIn{Reference image $\mathcal{I}^{\text{ref}}$, target viewpoints $\mathcal{V}_{1:N}$}
\Parameter{Iterations $K$, \\
linearly increasing CFG scale $\left\{w_k\right\}_{k=1}^K$}

\KwOut{Novel view RGB images and normal maps $\hat{\mathcal{I}}_{1:N}$}

Initialize SMPL-X params $\left(\theta, \beta, \psi \right) \leftarrow \operatorname{Estimate}\left(\mathcal{I}^{\text{ref}}\right)$

\For {$k = 1, \ldots, K-1$} {
    \tcp{Forward pass: Generate novel views with SMPL-X guidance and CFG scale $w_k$}
    $\hat{\mathcal{I}}_{1:N}$ = $\mathcal{G}\left(\mathcal{I}^{\text{ref}}, \mathcal{M}\left(\theta, \beta, \psi \right), \mathcal{V}_{1:N}; w_k\right)$ \label{alg:nvs}
    \BlankLine

    \tcp{Backward pass: Refine SMPL-X supervised by generated novel views}
    Optimize $\left(\theta, \beta \right)$ with $\hat{\mathcal{I}}_{1:N}$ by minimizing Eq.~\ref{eq:refinement-loss}
    \BlankLine

}
\tcp{Generation with final refined SMPL-X}

\Return{$\hat{\mathcal{I}}_{1:N}$ = $\mathcal{G}\left(\mathcal{I}^{\mathrm{ref}}, \mathcal{M}\left(\theta, \beta, \psi \right), \mathcal{V}_{1:N}; w_K\right)$}

\end{algorithm}

\begin{figure*}[t]
\centering
\includegraphics[width=\textwidth]{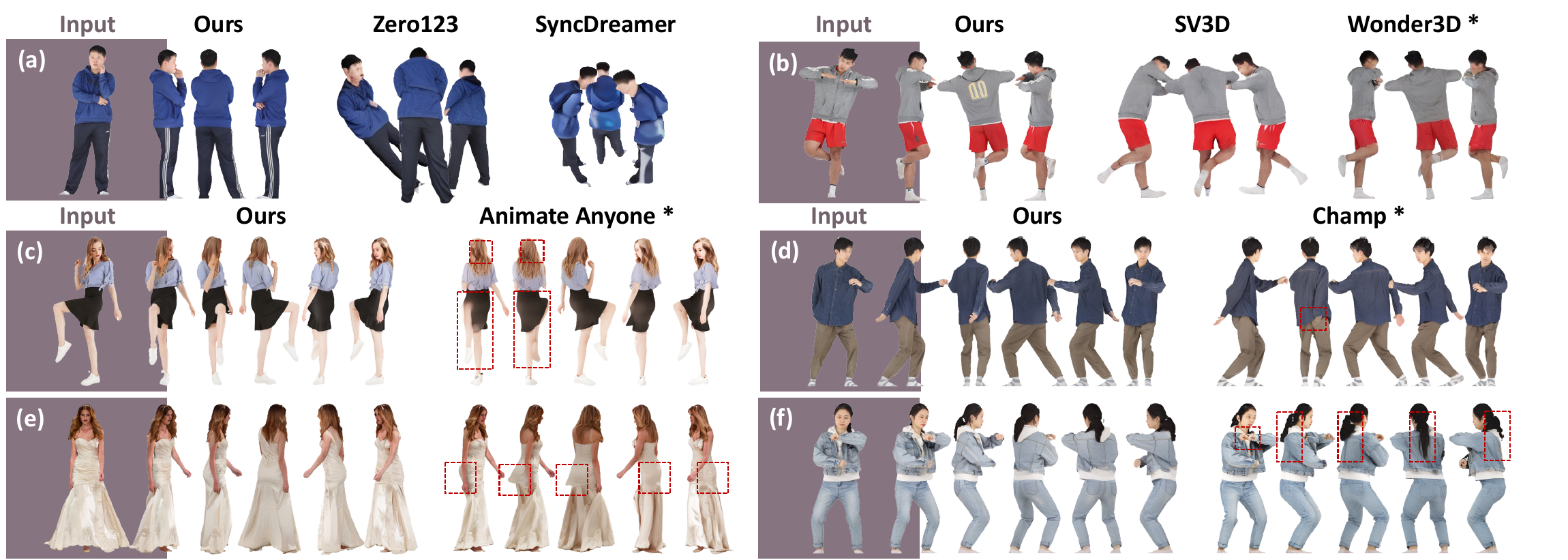}
\caption{\textbf{Qualitative results of novel view synthesis.} \methodname\ generates the highest-quality dense novel view images of humans with better consistency. * Methods are finetuned on THuman2.1 dataset for fair comparison. Please \textbf{zoom in} for details.}
\label{fig:comparison-nvs}
\end{figure*}

\subsection{Geometry-Aware Dual Branch}
\label{subsec:dual-branch}
Since the detailed geometry is difficult to capture within the RGB domain, we introduce the dual branch to perform geometry-aware denoising, which generates the spatially aligned normal maps along with RGB images. 
To be specific, We replicate just one input and output block of the UNet from the RGB branch to function as the normal branch, while the remaining blocks serve as shared components, as illustrated in Fig.~\ref{fig:pipeline}(b).
For the output of the normal branch, we use normal maps rendered from the ground-truth human scans as training supervision. For the input to the reference normal branch, considering the unavailability of the ground-truth during inference, we employ an off-the-shelf monocular normal estimator~\cite{ke2023repurposing} to estimate the reference normal map from the input RGB image, unifying the training and inference setting. 
With these designs, the shared blocks facilitate feature fusion across domains. The normal branch enriches RGB by incorporating geometry awareness, improving structural stability and geometric consistency, while RGB assists in generating more accurate and detailed normal maps. 
Notably, despite the input normal map inferred from the estimator typically being smooth and lacking details, our method could still produce the normal maps of output images with fine-grained details benefiting from the feature fusion across different views and domains.

\subsection{Iterative Refinement}
\label{subsec:method-optimization}
SMPL-X~\shortcite{pavlakos2019expressive} is a template human mesh $\mathcal{M}\left(\theta, \beta, \psi \right)$ parameterized by pose parameters $\theta$, shape parameters $\beta$, and expression parameters $\psi$.

The accuracy of the SMPL-X pose matters a lot since we employ its rendered normal and segmentation images as geometry guidance to improve 3D consistency. 
However, monocular estimation could give inaccurate SMPL-X poses that conflict with the reference images, leading to the generation of distorted novel view images and thus \emph{ill-shaped} 3D reconstruction as illustrated in Fig.~\ref{fig:ablation-refinement}(b).
On the other hand, generating novel view images without flawed SMPL-X guidance usually keeps its pose well matching the reference image, but inferior in terms of 3D consistency as shown in Fig.~\ref{fig:ablation-refinement}(a).
Therefore, we propose that multi-view human images generated without flawed SMPL-X guidance can be used to optimize the accuracy of SMPL-X poses, while the refined SMPL-X meshes can then guide the generation of human muti-views with improved 3D consistency.

Based on these observations, we randomly drop SMPL-X guidance with a certain ratio during training, enabling guidance-free generation in line with classifier-free guidance (CFG)~\shortcite{ho2022classifier}.
In the inference stage, we introduce an iterative refinement process, as detailed in Algorithm~\ref{alg:iterative-template-optimization-feedback}. Initially, we set the CFG scale to 0, effectively disabling SMPL-X guidance to preserve more accurate poses in the generated novel view images that match the reference image. These images are then used to update the SMPL-X parameters. In subsequent iterations, we gradually increase the CFG scale to enhance pose guidance of the refined SMPL-X estimation to further enhance 3D consistency.

Specifically, the iterative refinement process starts with estimating the initial SMPL-X parameters using PyMAF-X~\cite{pymaf2021,pymafx2023}. In each iteration, we use the current SMPL-X mesh as guidance, applying the corresponding CFG scale to generate human images. It's important to note that in the early iterations, the scale is kept small, resulting in weaker guidance that allows the generated images to better match the reference poses. Next, we use a differentiable renderer to produce SMPL-X mesh's normal maps $\mathcal{N}_{1:N}^{\text{SMPL-X}}$ and silhouettes $\mathcal{S}_{1:N}^{\text{SMPL-X}}$, as well as project 3D joints into 2D keypoints $\mathcal{J}_{1:N}^{\text{SMPL-X}}$ according to the camera poses. 
SMPL-X parameters are then optimized under the supervision of all the generated novel view images with the generated normal maps $\hat{\mathcal{N}}_{1:N}$, silhouettes $\hat{\mathcal{S}}_{1:N}$, and detected 2D joint keypoints $\hat{\mathcal{J}}_{1:N}$ from the generated novel views.
The SMPL-X optimization is performed by minimizing the following loss:
\begin{equation}
\fontsize{8}{12}
\begin{gathered}
\mathcal{L}_{\text{refine}} = \lambda_{\text{normal}}\mathcal{L}_{\text{normal}} + \lambda_{\text{silhouette}}\mathcal{L}_{\text{silhouette}} + \lambda_{\text{joint}}\mathcal{L}_{\text{joint}}, \\
\mathcal{L}_{\text{normal}} = \left| \mathcal{N}_{1:N}^{\text{SMPL-X}}-\hat{\mathcal{N}}_{1:N} \right|, 
\mathcal{L}_{\text{silhouette}} = \left| \mathcal{S}_{1:N}^{\text{SMPL-X}}-\hat{\mathcal{S}}_{1:N} \right|, \\
\mathcal{L}_{\text{joint}} = \left| \mathcal{J}_{1:N}^{\text{SMPL-X}}-\hat{\mathcal{J}}_{1:N} \right|.
\label{eq:refinement-loss}
\end{gathered}
\end{equation}
After the optimization, SMPL-X parameters are refined to be more accurate and aligned with the reference image, and will be fed back into the generation process with an increased CFG scale in the next iteration.

In summary, during each iterative process, the SMPL-X parameters undergo refinement supervised by all generated multi-view images, and the multi-view generation is enhanced with the improved SMPL-X as guidance.

\section{Experiments}
\label{sec:exp}
\input{sec/tab_comparison_nvs}
\myparagraph{Training data.}
We train MagicMan on 2347 human scans from THuman2.1 dataset~\shortcite{tao2021function4d}. RGB and normal images are rendered using a weak perspective camera on 20 fixed viewpoints looking at the scan with evenly distributed azimuths from $0^\circ$ to $360^\circ$, at 512$\times$512 resolution.

\myparagraph{Evaluation data.}
We test on 95 scans from THuman2.1 dataset and 30 scans from CustomHumans dataset~\shortcite{ho2023learning} and also evaluate on in-the-wild images, comprising 100 from SHHQ dataset~\cite{fu2022stylegan} and 120 from the Internet featuring varied poses, outfits, and styles.

\myparagraph{Metrics.}
Evaluation is conducted on two tasks: 1) Novel view synthesis. We use PSNR, SSIM, LPIPS, and CLIP scores to compare generated views w.r.t. the ground-truth images of corresponding views. For in-the-wild data, we calculate LPIPS of the generated reference view and CLIP scores of generated novel views w.r.t. the input image. 2) 3D human reconstruction. Following~\citeauthor{xiu2022icon}, we calculate Chamfer and P2S distance, and L2 normal errors (NE).

\begin{figure}[t]
\centering
\includegraphics[width=\linewidth]{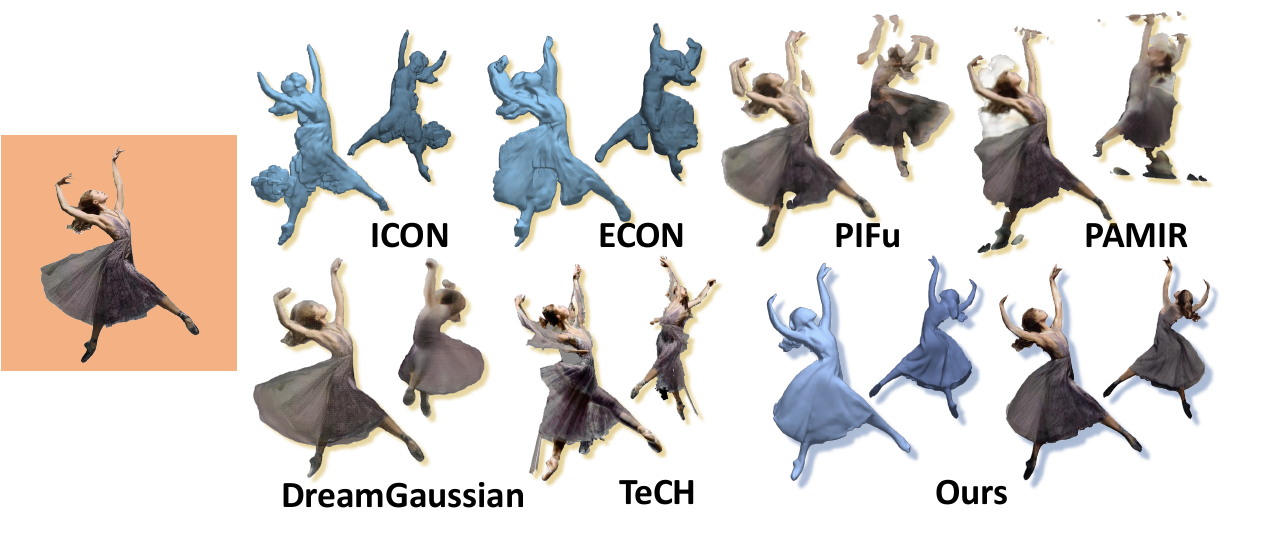}
\caption{\textbf{Reconstructed 3D human meshes.} \methodname\ produces the best geometry and textures in the case with challenging poses and loose outfits. Please \textbf{zoom in} for details.}
\label{fig:comparison-rec}
\end{figure}

\subsection{Novel View Synthesis}
To evaluate novel view synthesis results, we compare \methodname\ with generative object novel view synthesis methods, i.e. Zero123~\shortcite{liu2023zero}, SyncDreamer~\shortcite{liu2023syncdreamer}, Wonder3D~\shortcite{long2024wonder3d}, and SV3D~\shortcite{voleti2024sv3d}, and character animation methods with body priors, i.e. Animate Anyone~\shortcite{hu2024animate} and Champ~\shortcite{zhu2024champ}. 
Examples of human novel view images and normal maps generated by \methodname\ are shown in Fig.\ref{fig:teaser}, demonstrating that \methodname\ can generate high-quality and 3D consistent human novel views across diverse poses, outfits, and styles.
Fig.~\ref{fig:comparison-nvs} presents qualitative comparisons between \methodname\ and baseline methods. 
Zero123, SyncDreamer, and SV3D without finetuning typically generate distorted human images, suggesting that these methods are not suitable to be directly used as 3D priors for tasks concerning humans.
Wonder3D produces only six views at half of our resolution, leading to texture detail loss. Lack of a body prior also results in geometric errors.
Animation methods yield unrealistic body structures for a lack of geometric awareness, sometimes encountering ambiguities between front and back views as shown in Fig.~\ref{fig:comparison-nvs}(c). Besides, they exhibit noticeable inconsistency between views with a large viewpoint change as illustrated in Fig.~\ref{fig:comparison-nvs}(e) and~\ref{fig:comparison-nvs}(f).
In contrast, our method achieves stable structures, consistent geometry, and textures while generating dense novel views for humans. Further examples are provided in \suppl

Quantitative comparisons are reported in Tab~\ref{tab:comparison-nvs}. Results show that \methodname\ outperforms baseline approaches in both pixel-level and semantic metrics, except for slightly higher LPIPS on in-the-wild data for reference view reconstruction, likely due to SV3D's better frontal details at a higher resolution. However, CLIP scores of novel views indicate that our method significantly excels in novel view synthesis.

\subsection{3D Human Reconstruction}

\input{sec/tab_comparison_recon}

Fig.~\ref{fig:comparison-rec} displays our reconstructed human mesh, compared with those produced by baseline methods including feed-forward approaches PIFu~\shortcite{saito2019pifu}, PaMIR~\shortcite{zheng2021pamir}, ICON~\shortcite{xiu2022icon}, ECON~\shortcite{xiu2023econ}, and SDS-based DreamGaussian~\shortcite{tang2023dreamgaussian}, TeCH~\shortcite{huang2024tech}. Both feed-forward and SDS-based methods fail to produce reasonable geometry and detailed consistent textures for the challenging pose and outfit, while our 3D-aware diffusion model with refined body prior generates dense and consistent multi-views, which support reliable reconstruction with enhanced geometry and textures. Additional visual results are available in \suppl\ Quantitative comparisons with PIFu, PAMIR, ICON, and ECON are presented in Tab. \ref{tab:comparison-recon}, illustrating that \methodname\ outperforms previous approaches on all metrics by a significant margin. Note that we include our iterative refinement process and the SMPL-X optimization operations of ICON, ECON, and PAMIR are retained for fair comparison.

\subsection{Ablations and Discussions}
\input{sec/tab_ablation}
\begin{figure}[t]
\centering
\includegraphics[width=\linewidth]{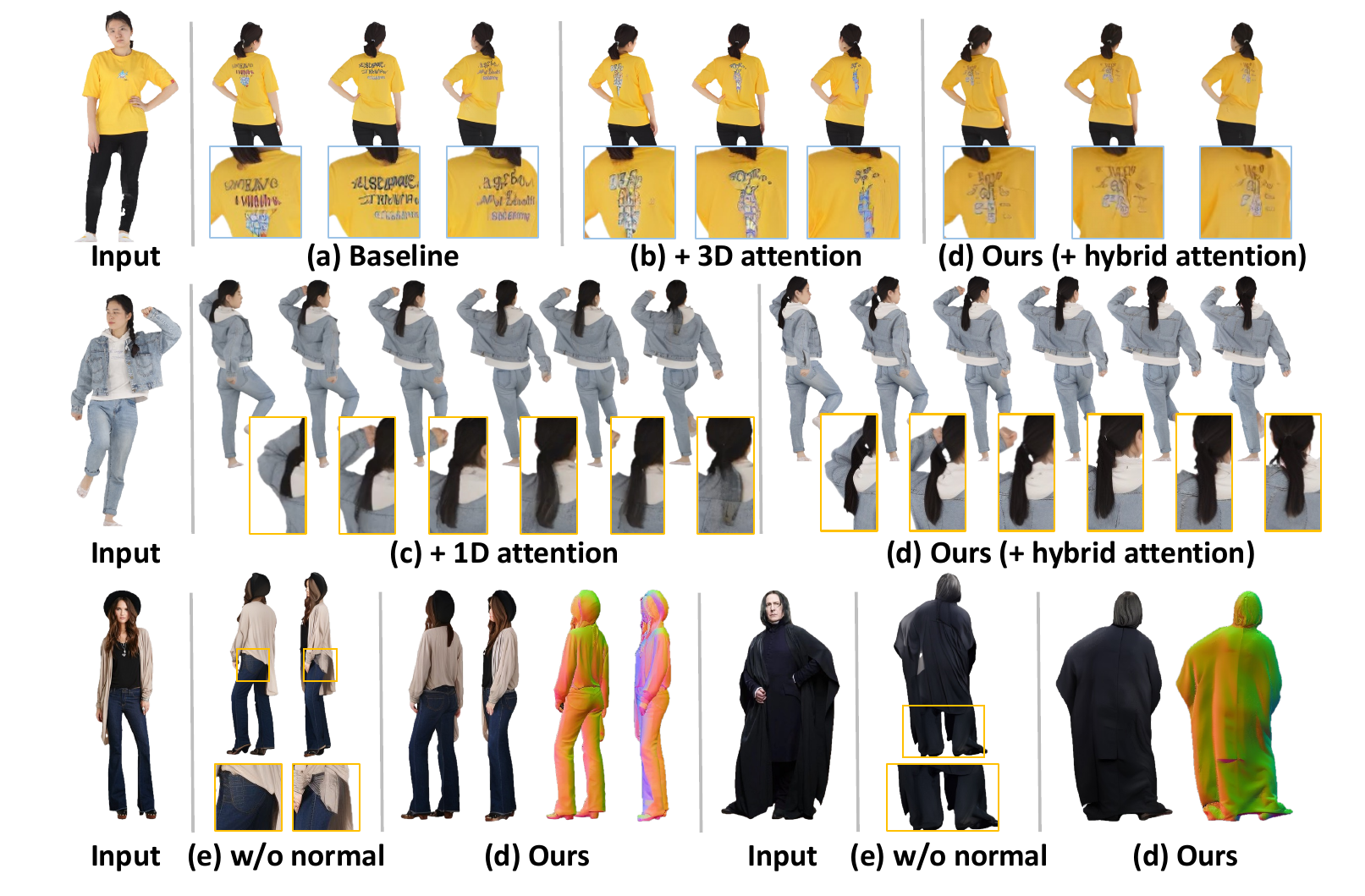}
\caption{\textbf{Ablations on hybrid attention and dual branch.} Our full model presents the best multi-view consistency.}
\label{fig:ablation-nvs}
\end{figure}

\myparagraph{Hybrid attention.} With hybrid attention, \methodname\ can generate up to 20 consistent multi-views in training, with an inference time of $\sim$40s on 1 A100 GPU, while traditional 3D attention across all views yields only 6 views under the same memory constraint and takes $\sim$60s for inference.
Fig.~\ref{fig:ablation-nvs} illustrates the effectiveness of different components of hybrid attention: (a) The baseline model without multi-view attention generates inconsistent views. (b) 3D attention across selected views still produces flickering cloth patterns. (c) 1D attention alone presents content drift, e.g., hair length that gradually changes with increasing viewpoint changes, indicating that information exchange via 1D attention alone improves similarity but is insufficient for comprehensive consistency.
(d) Our full model with hybrid attention demonstrates the best consistency when generating dense multi-views, also confirmed by the quantitative results in Tab.~\ref{tab:ablation}.

\myparagraph{Geometry-aware dual branch.} 
In Fig.~\ref{fig:ablation-nvs}(e) and row 4 of Tab.~\ref{tab:ablation}, removing the normal branch leads to a degradation in multi-view consistency, especially in complex geometric deformations, e.g., fabric layers and folds. Our full model with normal prediction enhances geometry awareness, yielding improved structures and consistency.
 
\begin{figure}[t]
\centering
\includegraphics[width=\linewidth]{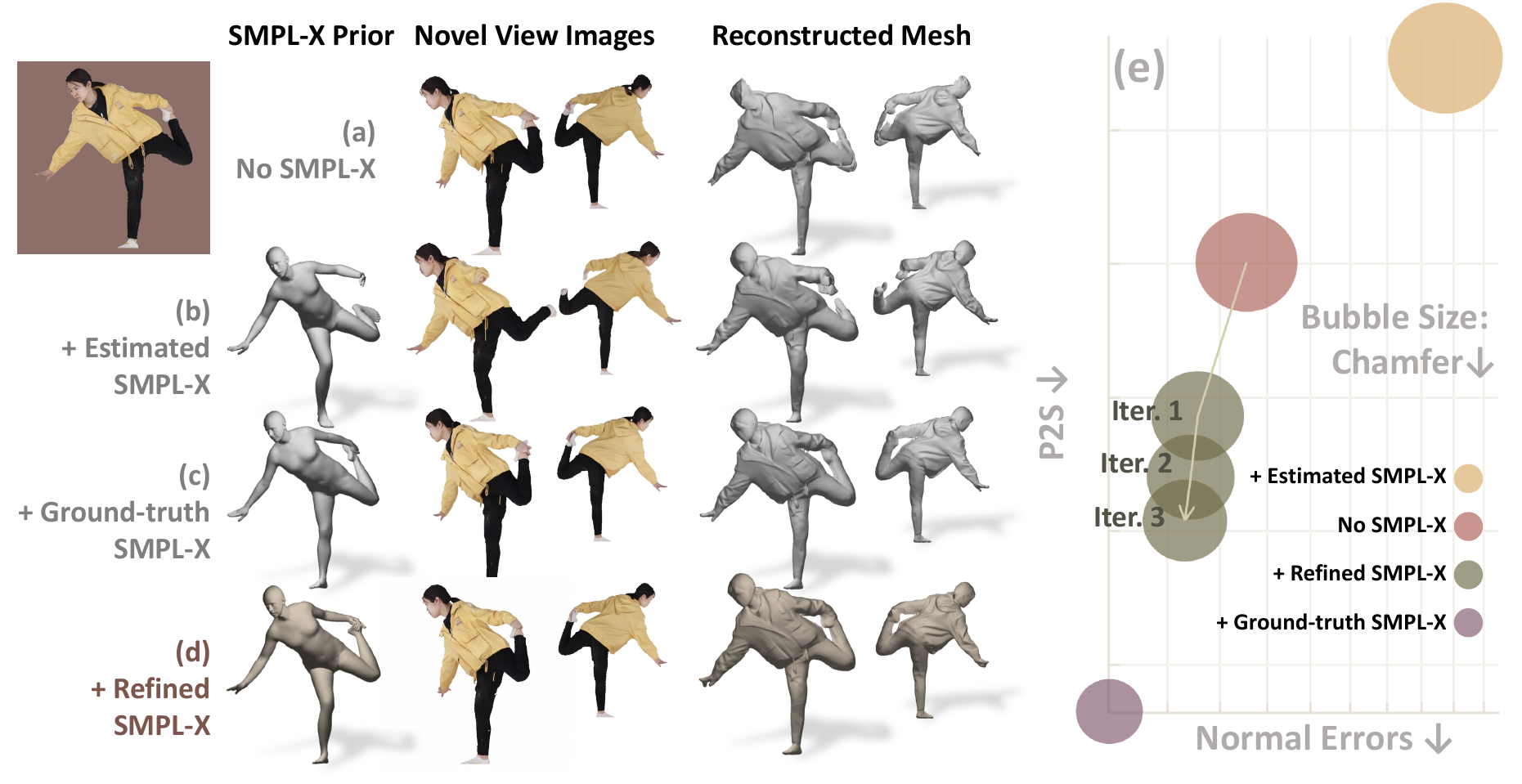}
\caption{\textbf{Qualitative (left) and quantitative (right) results of ablation studies on the iterative refinement.} Metrics are calculated on 50 scans from THuman2.1 dataset.}
\label{fig:ablation-refinement}
\end{figure}

\myparagraph{Iterative refinement.}
We conduct an ablation study to valid the effectiveness of the iterative refinement process.
As shown in Fig.~\ref{fig:ablation-refinement}(a), generation without SMPL-X guidance produces seemingly satisfactory novel views with accurate poses, which, however, exhibits severe artifacts in reconstruction due to inconsistent poses between views without 3D prior.
Directly utilizing the estimated inaccurate SMPL-X mesh as pose guidance in Fig.~\ref{fig:ablation-refinement}(b) leads to distorted novel view images and the \emph{ill-shaped} reconstructed mesh (e.g., the missing and disjointed hand and foot) due to conflicts between incorrect SMPL-X and reference image.
Impressive results can be achieved with the accurate ground-truth SMPL-X as presented in Fig.~\ref{fig:ablation-refinement}(c), which, however, is unavailable in practice.
Our iterative refinement process progressively improves the novel view results for reconstruction with increasingly accurate SMPL-X guidance through successive iterations, as demonstrated by the green bubbles in Fig.~\ref{fig:ablation-refinement}(e). The finally refined multi-view images, encompassing both accurate poses and 3D consistency, yield comparable results to those produced with the ground-truth SMPL-X.
The refined SMPL-X mesh with more accurate poses and reduced depth ambiguities, as a byproduct of our refinement process, suggests that the abundant priors in pre-trained image diffusion models can potentially aid in human body estimation.

Unlike previous optimization methods as introduced by SMPLify~\shortcite{pavlakos2019expressive}, ICON~\shortcite{xiu2022icon}, etc., which essentially aligned the SMPL-X mesh with the input image on a single-view 2D plane, our method fully leverages the multi-view outputs from the 3D-aware diffusion model to align the SMPL-X mesh with 3D geometry information, achieving more accurate pose in 3D space.

\section{Conclusion}
\label{sec:conclusion}
In this paper, we introduce \textbf{\methodname}, a method for generating human novel views from a single reference image by leveraging an image diffusion model as the 2D generative prior and the SMPL-X model as the 3D body prior. Building on this, our proposed efficient hybrid multi-view attention ensures the generation of denser multi-view images while maintaining high 3D consistency, which is further enhanced by the geometry-aware dual branch via geometry cues.
Moreover, our novel iterative refinement process optimizes the initially estimated SMPL-X poses over successive iterations, guiding novel view generation with improved consistency and alleviating \emph{ill-shaped} issues caused by inaccurate SMPL-X estimates. Extensive experimental results demonstrate that our method can generate dense, high-quality, and consistent human novel view images, which are also ideally suited for subsequent 3D human reconstruction tasks.

%% file: sec/tab_comparison_nvs.tex
\begin{table*}[t]
\centering
\setlength{\tabcolsep}{0.9mm} 
{\fontsize{9pt}{11pt}\selectfont
\begin{tabular}{@{}lcccccccccc@{}}
\toprule
\multicolumn{1}{c|}{}        & \multicolumn{4}{c|}{THuman2.1}                                             & \multicolumn{4}{c|}{CustomHumans}                                          & \multicolumn{2}{c}{in-the-wild} \\ 
\multicolumn{1}{l|}{Method}         & PSNR $\uparrow$        & SSIM $\uparrow$        & LPIPS $\downarrow$       & \multicolumn{1}{c|}{CLIP $\uparrow$}        & PSNR $\uparrow$        & SSIM $\uparrow$        & LPIPS $\downarrow$       & \multicolumn{1}{c|}{CLIP $\uparrow$}        & LPIPS $\downarrow$      & CLIP $\uparrow$               \\ \toprule
\multicolumn{1}{l|}{Zero123}        & 15.5/12.3 & 0.829/0.798 & 0.215/0.247 & \multicolumn{1}{c|}{0.792/0.773} & 13.9/10.3 & 0.811/0.766 & 0.211/0.250 & \multicolumn{1}{c|}{0.818/0.800} & 0.082      & 0.722              \\
\multicolumn{1}{l|}{SyncDreamer}    & 13.2/\phantom{1}--\phantom{.3} & 0.841/\phantom{0.}--\phantom{98} & 0.209/\phantom{0.}--\phantom{47} & \multicolumn{1}{c|}{0.749/\phantom{0.}--\phantom{73}} & 11.5/\phantom{1}--\phantom{.0} & 0.824/\phantom{0.}--\phantom{66} & 0.211/\phantom{0.}--\phantom{50} & \multicolumn{1}{c|}{0.767/\phantom{0.}--\phantom{00}} & 0.175      & 0.630              \\
\multicolumn{1}{l|}{SV3D}           & 19.8/16.5 & 0.896/0.868 & 0.115/0.140 & \multicolumn{1}{c|}{0.888/0.877} & 18.3/14.3 & 0.892/0.857 & 0.111/0.141 & \multicolumn{1}{c|}{0.912/0.899} & \textbf{0.030}      & 0.818              \\
\multicolumn{1}{l|}{Wonder3D*}      & 21.2/\phantom{1}--\phantom{.3} & 0.906/\phantom{0.}--\phantom{98} & 0.110/\phantom{0.}--\phantom{47} & \multicolumn{1}{c|}{0.882/\phantom{0.}--\phantom{734}} & 18.3/\phantom{1}--\phantom{.0} & 0.889/\phantom{0.}--\phantom{66} & 0.122/\phantom{0.}--\phantom{50} & \multicolumn{1}{c|}{0.863/\phantom{0.}--\phantom{00}} & 0.064      & 0.757              \\ \midrule
\multicolumn{1}{l|}{Animate Anyone*} & 23.6/22.1 & 0.923/0.910 & 0.070/0.078 & \multicolumn{1}{c|}{0.920/0.917} & 22.0/20.4 & 0.919/0.905 & 0.060/0.068 & \multicolumn{1}{c|}{0.931/0.929} & 0.061      & 0.850              \\
\multicolumn{1}{l|}{Champ*}          & \underline{24.9}/\underline{23.3} & \underline{0.930}/\underline{0.918} & \underline{0.063}/\underline{0.071} & \multicolumn{1}{c|}{\underline{0.927}/\underline{0.924}} & \underline{23.2}/\underline{21.4} & \underline{0.931}/\underline{0.918} & \underline{0.055}/\underline{0.063} & \multicolumn{1}{c|}{\underline{0.938}/\underline{0.934}} & 0.053      & \underline{0.852}              \\ \midrule[0.6pt]
\multicolumn{1}{l|}{Ours}           & \textbf{26.0}/\textbf{24.9} & \textbf{0.946}/\textbf{0.929} & \textbf{0.049}/\textbf{0.054} & \multicolumn{1}{c|}{\textbf{0.947}/\textbf{0.938}} & \textbf{24.7}/\textbf{22.9} & \textbf{0.950}/\textbf{0.937} & \textbf{0.044}/\textbf{0.052} & \multicolumn{1}{c|}{\textbf{0.947}/\textbf{0.947}} & \underline{0.040}      & \textbf{0.871}     \\ \bottomrule
\end{tabular}
}
\caption{\textbf{Quantitative evaluation for novel view synthesis on ``4 views / 20 views''.} * Methods are finetuned on THuman2.1 dataset for fair comparison. Top two are highlighted as \textbf{first} and \underline{second}. Views selection for evaluation is detailed in \suppl}
\label{tab:comparison-nvs}
\end{table*}

%% file: sec/tab_comparison_recon.tex
\begin{table}[]
\centering
\setlength{\tabcolsep}{1.4mm} 
{\fontsize{9pt}{11pt}\selectfont
\begin{tabular}{@{}lcccccc@{}}
\toprule
      & \multicolumn{3}{c}{THuman2.1}         & \multicolumn{3}{c}{CustomHumans}      \\ \cmidrule(lr){2-4} \cmidrule(lr){5-7} 
Method & Chamfer $\downarrow$ & P2S $\downarrow$ & NE $\downarrow$ & Chamfer $\downarrow$ & P2S $\downarrow$ & NE $\downarrow$ \\ \midrule
PIFu  & 5.62       & 5.11       & 0.150       & 6.43       & 5.76       & 0.159       \\
PAMIR & {\ul 4.30} & {\ul 4.27} & {\ul 0.132} & {\ul 5.00} & {\ul 4.89} & 0.135       \\
ICON  & 5.05       & 5.02       & 0.139       & 5.46       & 5.48       & {\ul 0.134} \\
ECON  & 5.45       & 5.26       & 0.148       & 5.72       & 5.61       & 0.138       \\ \midrule
Ours   & \textbf{2.35}        & \textbf{2.44}    & \textbf{0.093}      & \textbf{2.34}        & \textbf{2.43}    & \textbf{0.095}      \\ \bottomrule
\end{tabular}
}
\caption{\textbf{Quantitative evaluation for 3D human reconstruction.} Top two are highlighted as \textbf{first} and \underline{second}.}
\label{tab:comparison-recon}
\end{table}

%% file: sec/tab_ablation.tex
\begin{table}[t]
\centering
\setlength{\tabcolsep}{0.8mm} 
{\fontsize{9pt}{11pt}\selectfont
\begin{tabular}{@{}cl|cccccc@{}}
\toprule
\multicolumn{2}{c}{}    & \multicolumn{4}{c}{THuman2.1}                                           & \multicolumn{2}{c}{in-the-wild}      \\  \cmidrule(lr){3-6}  \cmidrule(lr){7-8} 
\multicolumn{2}{c}{Method}     & PSNR $\uparrow$ & SSIM $\uparrow$ & LPIPS $\downarrow$ & CLIP $\uparrow$ & LPIPS $\downarrow$ & CLIP $\uparrow$ \\ \midrule
\multicolumn{2}{c}{Baseline}   & 22.84           & 0.915           & 0.070              & 0.923           & \textbf{0.037}     & 0.843           \\
\multicolumn{2}{c}{+ 1D attn.} & {\ul 23.94}     & {\ul 0.924}     & 0.063              & 0.935           & 0.041              & 0.853           \\
\multicolumn{2}{c}{+ 3D attn.} & 23.77           & 0.923           & {\ul 0.063}        & {\ul 0.936}     & 0.042              & {\ul 0.854}     \\ \midrule
\multicolumn{2}{c}{w/o normal}   & 23.73           & 0.921           & 0.064              & 0.925           & 0.042              & 0.833           \\ \midrule
\multicolumn{2}{c}{Ours}       & \textbf{24.85}  & \textbf{0.929}  & \textbf{0.054}     & \textbf{0.938}  & {\ul 0.040}        & \textbf{0.871}  \\ \bottomrule
\end{tabular}
}
\caption{\textbf{Quantitative ablation results on hybrid attention and dual branch.} Top two are highlighted as \textbf{first} and \underline{second}.}
\label{tab:ablation}
\end{table}

%% file: sec/supp.tex
\clearpage
\setcounter{page}{1}
\setcounter{section}{0}
\setcounter{figure}{0}
\setcounter{table}{0}
\setcounter{equation}{0}
\onecolumn
\begin{center}
\LARGE{\textbf{\thetitle}}\\ \vspace{0.5em}
\end{center}
\section*{Supplementary Material}
\label{sec:supp}
\vspace{0.5em}
\renewcommand\thesection{\Alph{section}}
\renewcommand\thefigure{\Alph{section}\arabic{figure}}
\renewcommand\thetable{\Alph{section}\arabic{table}}
\renewcommand\theequation{\Alph{section}\arabic{equation}}

In this supplementary document, we will introduce the following contents: 
\begin{itemize}
    \item More discussion about the iterative refinement (Sec.~\ref{sec:supp-iterative-refinement});
    \item Mehtod and experiment details, including details of the conditional diffusion model (Sec.~\ref{sec:supp-conditional-diffusion}), hybrid multi-view attention (Sec.~\ref{sec:supp-hybrid-attention}), geometry-aware dual branch (Sec.~\ref{sec:supp-dual-branch}), and other implementation details (Sec.~\ref{sec:supp-other-details});
    \item Discussion about limitations and future works (Sec.~\ref{sec:supp-limitations}), and ethical consideration (Sec.~\ref{sec:supp-ethical})
    \item More qualitative results (Sec.~\ref{sec:supp-results}).
\end{itemize}

Given the inclusion of more mathematical expressions, we have opted for a single-column format in this supplementary document to enhance readability. More visual videos, code, and other resources can be found at \projectpage.

\section{Discussion of Iterative Refinement}
\label{sec:supp-iterative-refinement}
\myparagraph{\emph{Ill-shaped} problem caused by inaccurate SMPL-X.} 
Existing works in human reconstruction and generation~\cite{xiu2022icon,xiu2023econ,zheng2021pamir,liu2024humangaussian,huang2024tech} highlight the importance of using parametric human models, such as SMPL-X in our case, to enhance generalization and robustness. However, in practice, SMPL-X meshes from monocular estimation are often inaccurate, featuring misalignment with the reference image and depth ambiguities. Using such flawed SMPL-X meshes for generation and reconstruction can significantly degrade quality, particularly causing issues in geometry, which we refer to as \emph{ill-shaped} problems. 

\begin{figure*}[t]
\centering
\includegraphics[width=0.86\textwidth]{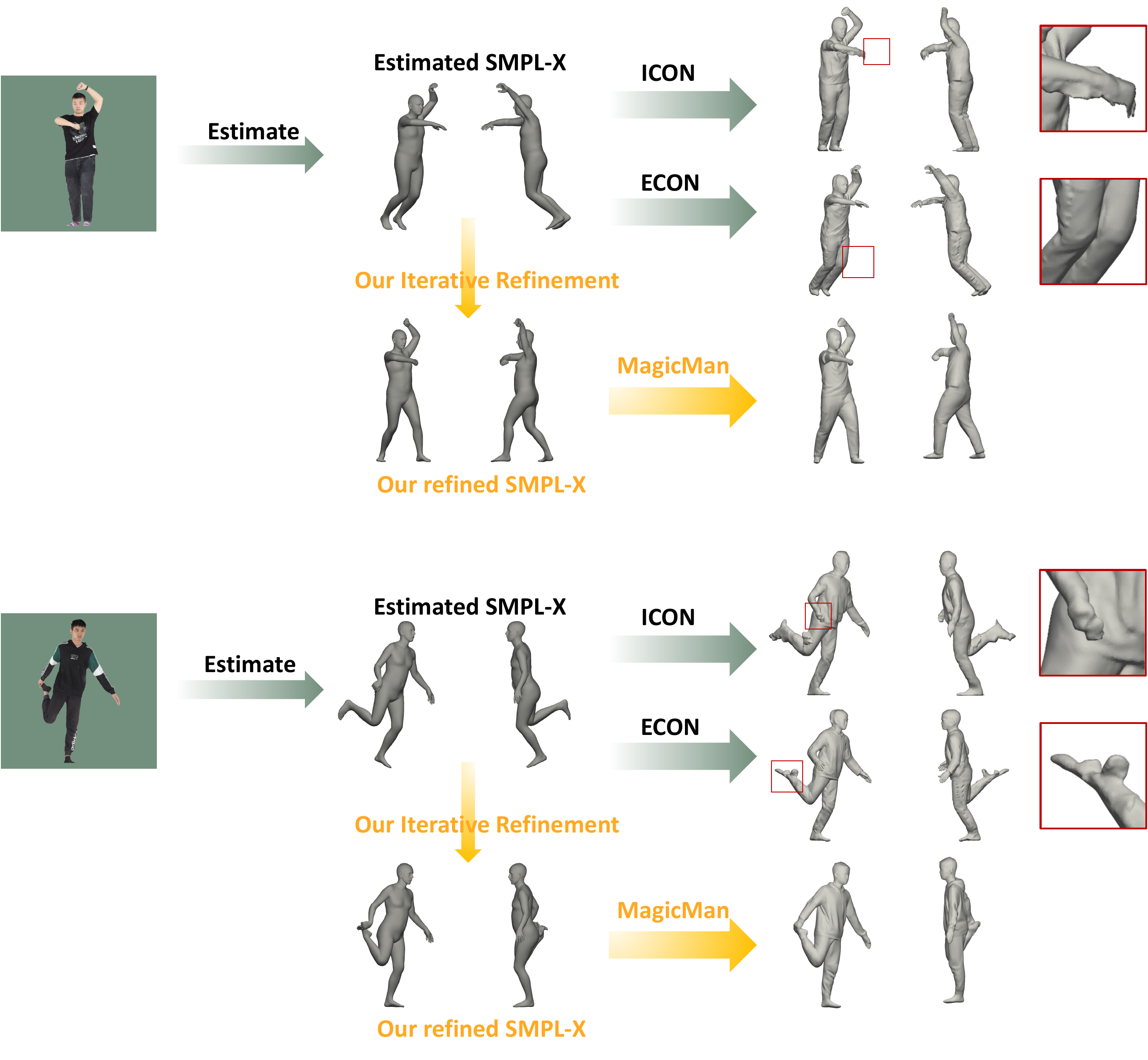}
\caption{\textbf{Typical \emph{ill-shaped} problems caused by inaccurate SMPL-X poses} in 3D human reconstruction: 1) abnormal overall poses like tilted body (upper); 2) broken and distorted body parts (lower). Our proposed iterative refinement significantly improves the accuracy of SMPL-X poses, alleviating \emph{ill-shaped} problems. Please \faSearch\ \textbf{zoom in} for details.
}
\label{fig:supp-ill}
\end{figure*}

In Fig.~\ref{fig:supp-ill}, we illustrate common \emph{ill-shaped} problems that arise when inaccurate SMPL-X meshes are used in ICON~\shortcite{xiu2022icon} and ECON~\shortcite{xiu2023econ} for 3D human reconstruction. 
There are two typical types of \emph{ill-shaped} problems: 1) abnormal overall poses due to depth ambiguities, such as tilted and floating bodies or bent knees observed from side views (as shown in the upper part of Fig.~\ref{fig:supp-ill}); 2) deformed or distorted body parts from mismatches between SMPL-X poses and the reference image, such as broken and incorrectly separated hands and feet (as shown in the lower part of Fig.~\ref{fig:supp-ill}).

Through our iterative refinement with generated multi-view human images, more accurate SMPL-X poses are achieved and significantly reduce the \emph{ill-shaped} problems in the reconstruction results.

\myparagraph{Difference from other SMPL-X optimization methods.} 
Actually, ICON~\shortcite{xiu2022icon} and ECON~\shortcite{xiu2023econ} also include optimization procedures for SMPL-X, which attempt to align the SMPL-X mesh with the predicted front and back normal maps. These optimizations are retained for the comparisons of ICON and ECON in this paper. Additionally, similar operations are also utilized in other monocular pose estimation methods to further align SMPL-X with the input image like SMPLify-X~\cite{pavlakos2019expressive}. However, this optimization approach fundamentally performs alignment in 2D space, without addressing the inherent limitations of monocular estimation, i.e., the lack of 3D information or depth information.

In our iterative optimization, we leverage the abundant human image priors in diffusion models to generate multi-view human images and thus can perform SMPL-X alignment in 3D space. This process introduces additional depth information relative to the reference image, transforming the ill-conditioned monocular SMPL-X estimation into a better-conditioned multi-view estimation problem to some extent.

\myparagraph{Implementation details.}
For iterative refinement, we set the number of iterative loops, $K$, to 4, and the linearly increasing CFG scales to $\left\{w_k\right\}_{k=1}^K = [0.0, 1.0, 2.0, 3.0 ]$, where $0.0$ corresponds to guidance-free generation initially, and $3.0$ to the finally refined novel view outputs. In the initial 3 iterations, the DDIM~\cite{song2020denoising} denoising steps for novel view generation are set to 15 to facilitate rapid iteration. In the final iteration, the steps are increased to 25 for higher-quality novel view images. For the optimization loss of SMPL-X parameters, we set $\lambda_{normal}=0.5, \lambda_{silh}=1.0, \lambda_{joint}=50.0$ for Eq. 3 in the main paper, and optimization steps to $[40, 20, 10]$ in the initial three iterations. The entire iterative refinement process takes $\sim$5 minutes on 1 A100 GPU.

\myparagraph{Additional Ablation results.} Corresponding to the bubble chart in Fig. 6 (right) of the main paper, we conduct ablation studies to validate the effectiveness of our proposed iterative refinement, with the specific numerical results reported in Tab~\ref{tab:supp-refinement}. Note the following: 1) Since SMPL-X and the refinement process primarily impact pose accuracy and 3D geometric consistency, which are not readily apparent through novel view generation results, we rely on subsequent reconstruction outcomes to assess the quality of generated novel view images. 2) Considering the time required to independently optimize and reconstruct the mesh for each iteration in this evaluation, we report the results of randomly selected 50 scans from the Thuman2.1 dataset instead of the full testing set. 3) Results with the ground-truth SMPL-X guidance are presented as the upper limit in the last row, which are, in fact, unavailable in practice.

Results show that: 1) Compared to directly using estimated SMPL-X meshes with inaccurate poses as guidance, iterative refinement significantly improves the accuracy of SMPL-X poses, reduces conflicts between pose guidance and the reference image, and finally alleviates \emph{ill-shaped} issues like tilted bodies and deformed body parts in both generation and reconstruction results. 2) During the iterative loops, optimization starts without SMPL-X guidance. As the SMPL-X guidance becomes progressively more accurate, it enhances the 3D geometric consistency of the generated novel view images, thereby improving the final reconstruction quality.

\begin{table}[t]
\centering
\begin{tabular}{@{}llccc@{}}
\toprule
\multicolumn{2}{l}{Method}      & Chamfer $\downarrow$ & P2S $\downarrow$ & Normal $\downarrow$ \\ \midrule
+ Esimated SMPL-X    &         & 3.73                 & 3.77             & 0.157               \\ \midrule
w/o SMPL-X             &         & 2.96                 & 3.01             & 0.106               \\
                      & Iteration 1 & 2.43                 & 2.43             & 0.096               \\
                      & Iteration 2 & {\ul 2.19}           & {\ul 2.20}       & {\ul 0.094}         \\
\multirow{-3}{*}{+ Refined SMPL-X} &
  \cellcolor[HTML]{EFEFEF}Iteration 3 &
  \cellcolor[HTML]{EFEFEF}\textbf{2.03} &
  \cellcolor[HTML]{EFEFEF}\textbf{2.04} &
  \cellcolor[HTML]{EFEFEF}\textbf{0.093} \\ \midrule
+ Ground-truth SMPL-X &         & 1.27                 & 1.33             & 0.080               \\ \bottomrule
\end{tabular}
\caption{\textbf{Qualitative ablation result of iterative refinement} on 50 scans of THuman2.1 dataset. Top two are highlighted as \textbf{first} and \underline{second}.}
\label{tab:supp-refinement}
\end{table}

\section{Method and Experiment Details}
\label{sec:supp-details}
\subsection{Details of Conditional Diffusion Model}
\label{sec:supp-conditional-diffusion}
\myparagraph{Camera embeddings.}
Similar to zero123~\cite{liu2023zero}, we use a spherical coordinate system to represent relative viewpoint transformations. Given the center of the human scan is in the center of the coordinate system, we represent the elevation, azimuth, and radius as $\theta$, $\phi$, and $r$, respectively, and the camera always looks at the center of the scan. Then a generated viewpoint is calculated relative to the reference viewpoint as $\left(\Delta \theta, \Delta \phi, \Delta r \right) = \left(\theta_i-\theta_\text{ref}, \phi_i-\phi_\text{ref}, r_i-r_\text{ref} \right)$. Under our weak perspective camera and uniformly distributed azimuthal orbit viewpoints, the generated viewpoint shares the same $\theta$ and $r$ as the reference viewpoint, i.e., $\left(\Delta \theta, \Delta \phi, \Delta r \right) = (0, \phi_i-\phi_\text{ref}, 0)$. This relative viewpoint is employed to compute the cam2world transformation matrix and can be fully determined by the rotation matrix $R \in \mathbb{R}^{3\times3}$. The rotation matrix $R$ is flattened into a vector of $\mathbb{R}^{9}$ and subsequently encoded by an MLP into camera embeddings of $\mathbb{R}^{1024}$, which is then added to the denoising time embeddings as the viewpoint control.

\subsection{Details of Hybrid Attention}
\label{sec:supp-hybrid-attention}
\begin{figure*}[t]
\centering
\includegraphics[width=\textwidth]{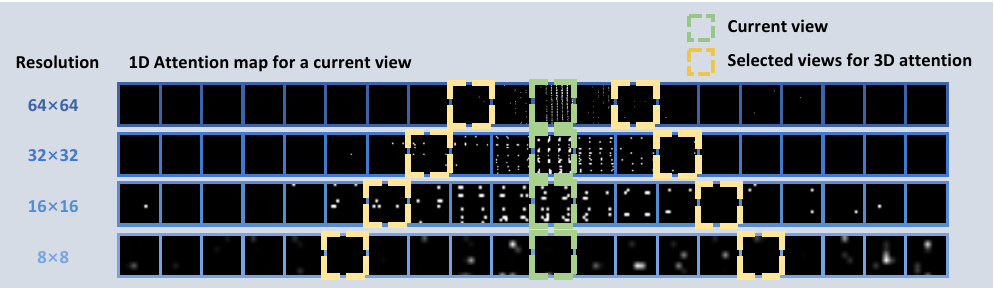}
\caption{\textbf{Attention map for 1D attention} and \textbf{the schematic diagram for view selection of 3D attention}. For clarity in presentation, all attention maps at different resolutions (as indicated on the left) are normalized to the same size, with adjacent feature maps having an $18^\circ$ angular difference in our experimental setup.
}
\label{fig:supp-attention}
\end{figure*}

\myparagraph{View selection strategy for 3D attention.} As introduced in the main paper, our 3D attention operation is restricted to a sparse subset of views to reduce computational costs, with the selection varying across denoising UNet blocks to fully leverage different levels of information. For the view selection strategy, we base our solution on the observation that, UNet blocks at higher resolution, closer to the input and output, handle more concrete low-level information, like detailed textures, which is more relevant to nearby views. Conversely, middle blocks at lower resolution process more abstract high-level information, such as body structures and poses, which can be relevant across more distant views. Here we use ``nearby'' and ``distant'' to describe the angular differences between views for the sake of brevity and clarity. Additionally, we observe that the range of views connected by 1D attention also follows the same pattern across different blocks, as shown in Fig.~\ref{fig:supp-attention}, where brighter areas in the attention map for 1D attention indicate higher attention values. 

Therefore, our overall approach is to apply 3D attention with nearby views at higher resolution and with more distant views at lower resolutions, and the specific view selection approach at different blocks is shown in Fig.~\ref{fig:supp-attention} and Tab.~\ref{tab:supp-view-selection} with an $18^\circ$ angular difference between two adjacent views.
In our experiments, we find that with the view selection strategy varying across block levels, 
the feature map of each view needs to exchange information with only 2 other views through 3D attention, instead of all 20 views in the original 3D multi-view attention, significantly reducing computational cost while maintaining impressive multi-view consistency and alleviating content drift issues caused by using 1D attention alone.

\myparagraph{Ablations on view selection strategies.}
We also compare our view selection strategy with an alternative, counter-intuitive \emph{reverse} selection strategy that chooses distant views at higher resolutions and nearby views at lower resolutions as detailed in the last column of Tab.~\ref{tab:supp-view-selection}. The results of both strategies, along with the outcome of using 1D attention alone on THuman2.1~\cite{tao2021function4d} and in-the-wild data, are reported in the Tab.~\ref{tab:supp-ablation-view}, showing that our strategy more effectively ensures consistency with the same number of selected views.

\begin{table}[t]
\centering
\begin{tabular}{@{}cccc@{}}
\toprule
UNet block  & Resolution   & Selected views (ours)  & Selected views (ablation) \\ \midrule
DownBlock 1 & 64$\times$64 & $\{-36^\circ, +36^\circ\}$ & $\{-90^\circ, +90^\circ\}$    \\
DownBlock 2 & 32$\times$32 & $\{-54^\circ, +54^\circ\}$ & $\{-72^\circ, +72^\circ\}$    \\
DownBlock 3 & 16$\times$16 & $\{-72^\circ, +72^\circ\}$ & $\{-54^\circ, +54^\circ\}$    \\
DownBlock 4 & 8$\times$8   & $\{-90^\circ, +90^\circ\}$ & $\{-36^\circ, +36^\circ\}$    \\
MiddleBlock & 8$\times$8   & $\{-90^\circ, +90^\circ\}$ & $\{-36^\circ, +36^\circ\}$    \\
UpBlock 1   & 16$\times$16 & $\{-90^\circ, +90^\circ\}$ & $\{-36^\circ, +36^\circ\}$    \\
UpBlock 2   & 32$\times$32 & $\{-72^\circ, +72^\circ\}$ & $\{-54^\circ, +54^\circ\}$    \\
UpBlock 3   & 64$\times$64 & $\{-54^\circ, +54^\circ\}$ & $\{-72^\circ, +72^\circ\}$    \\
UpBlock 4   & 64$\times$64 & $\{-36^\circ, +36^\circ\}$ & $\{-90^\circ, +90^\circ\}$    \\ \bottomrule
\end{tabular}
\caption{\textbf{View selection strategy for 3D attention across different UNet blocks.} In our experimental setup the different views are spaced at $18^\circ$ intervals, so we use angles to represent the selected views relative to the central current view in this table, e.g., $\{-36^\circ, +36^\circ\}$ indicates that for each view, the other two views with an azimuth difference of $\pm 36^\circ$ are selected to perform 3D attention. ``Selected views (ours)'' indicates the view selection strategy we ultimately used (Sec.~\ref{sec:supp-hybrid-attention}), while ``Selected views (ablation)'' represents an alternative \emph{reverse} strategy used for ablation studies.}
\label{tab:supp-view-selection}
\end{table}

\begin{table}[t]
\centering
\setlength{\tabcolsep}{0.85mm} 
{\fontsize{9pt}{11pt}\selectfont
\begin{tabular}{@{}cl|cccccc@{}}
\toprule
\multicolumn{2}{c}{}    & \multicolumn{4}{c}{THuman2.1}                                           & \multicolumn{2}{c}{in-the-wild}      \\  \cmidrule(lr){3-6}  \cmidrule(lr){7-8} 
\multicolumn{2}{l}{Method}     & PSNR $\uparrow$ & SSIM $\uparrow$ & LPIPS $\downarrow$ & CLIP $\uparrow$ & LPIPS $\downarrow$ & CLIP $\uparrow$ \\ \midrule
\multicolumn{2}{l}{1D attention} & {\ul 23.94}     & {\ul 0.924}     & 0.063              & 0.935           & 0.041              & 0.853           \\
\multicolumn{2}{l}{Hybird attention w/ reverse view selection} & 23.88     & 0.923     & {\ul 0.058}              & {\ul 0.937}           & {\ul 0.041}              & {\ul 0.860}           \\
\multicolumn{2}{l}{\cellcolor[HTML]{EFEFEF}Hybrid attention w/ our view selection}       & \cellcolor[HTML]{EFEFEF}\textbf{24.85}  & \cellcolor[HTML]{EFEFEF}\textbf{0.929}  & \cellcolor[HTML]{EFEFEF}\textbf{0.054}     & \cellcolor[HTML]{EFEFEF}\textbf{0.938}  & \cellcolor[HTML]{EFEFEF}\textbf{0.040}        & \cellcolor[HTML]{EFEFEF}\textbf{0.871}  \\ \bottomrule
\end{tabular}
}
\caption{\textbf{Quantitative ablation results on hybrid attention with different view selection strategies for 3D attention.} Top two are highlighted as \textbf{first} and \underline{second}. ``Hybird attention w/ reverse view selection'' corresponds to the last column of Tab.~\ref{tab:supp-view-selection}.}
\label{tab:supp-ablation-view}
\end{table}

\subsection{Details of Geometry-Aware Dual Branch}
\label{sec:supp-dual-branch}
Following~\citeauthor{liu2024humangaussian}, we replicate the denoising UNet's \texttt{conv\_in}, the first \texttt{DownBlock}, the last \texttt{UpBlock}, and \texttt{conv\_out} layers to create a normal branch identical to the RGB branch. The remaining blocks are shared between the two branches. The RGB branch is conditioned on the reference RGB image $\mathbf{x}_{\text{ref}}$ and SMPL-X segmentation map $\mathbf{p}_{\text{seg}}$ as pose guidance to generate RGB latents $\mathbf{x}$, while the normal branch is conditioned on the estimated reference normal map $\mathbf{n}_{\text{ref}}$ and SMPL-X normal map $\mathbf{p}_{\text{norm}}$ as pose guidance to generate normal latents $\mathbf{n}$. For the notation simplicity, we denote pixel-/latent-space conditions and targets with the same variable in this section. Normal latents are normalized to the similar distribution of RGB latents before being fed into the denoising UNet.
The $\mathbf{v}$-prediction~\cite{salimans2022progressive} learning target is employed as the network target for training the RGB and normal branches jointly:
\begin{equation}
\mathcal{L}^{\mathbf{v} \text {-pred }}=\mathbb{E}_{\mathbf{x}, \mathbf{n}, \mathbf{x}_{\text{ref}}, \mathbf{n}_{\text{ref}}, \mathbf{p}_{\text{seg}}, \mathbf{p}_{\text{norm}}, \mathbf{v}, t}\left[\left\|\hat{\mathbf{v}}_{\boldsymbol{\theta}}\left(\mathbf{x}_t ; \mathbf{x}_{\text{ref}}, \mathbf{p}_{\text{seg}}\right)-\mathbf{v}_t^{\mathbf{x}}\right\|_2^2+\left\|\hat{\mathbf{v}}_{\boldsymbol{\theta}}\left(\mathbf{n}_t ; \mathbf{n}_{\text{ref}}, \mathbf{p}_{\text{norm}}\right)-\mathbf{v}_t^{\mathbf{n}}\right\|_2^2\right],
\end{equation}
where $\mathbf{x}_t=\alpha_t \mathbf{x}+\sigma_t \boldsymbol{\epsilon}_{\mathbf{x}}$ and $\mathbf{n}_t=\alpha_t \mathbf{n}+\sigma_t \boldsymbol{\epsilon}_{\mathbf{n}}$ denote the noised RGB latents and normal latents, $\boldsymbol{\epsilon}_{\mathbf{x}}, \boldsymbol{\epsilon}_{\mathbf{n}} \sim \mathcal{N}(\mathbf{0}, \mathbf{I})$ independently sampled noise, $\mathbf{v}_t^{\mathbf{x}}=\alpha_t \boldsymbol{\epsilon}_{\mathbf{x}}-\sigma_t \mathbf{x}$ and $\mathbf{v}_t^{\mathbf{n}}=\alpha_t \boldsymbol{\epsilon}_{\mathbf{n}}-\sigma_t \mathbf{n}$ the $\mathbf{v}$-prediction learning targets at time step $t$ for RGB and normal respectively. 

\subsection{Data Processing and Other Implementation Details}
\label{sec:supp-other-details}
\myparagraph{Training data.}
We train MagicMan on 2347 human scans from THuman2.1 dataset~\cite{tao2021function4d}. All scans are scaled to occupy 90\% of the image height and are centered. Then we render RGB and normal images at the resolution of 512$\times$512 for training using a weak perspective camera on $N=20$ fixed viewpoints looking at the scan with evenly distributed azimuths from $0^\circ$ to $360^\circ$ and a random elevation between $\left[-15^\circ, 15^\circ\right]$. Front views with random azimuthal rotations between $\left[-18^\circ, 18^\circ\right]$ serve as the reference for better generalizability. 

\myparagraph{Evaluation data.}
We test on 95 scans from THuman2.1 dataset with identities unseen during training and 30 scans from CustomHumans dataset~\cite{ho2023learning}, and the front view is used as the reference. Moreover, we conduct evaluations on in-the-wild images, comprising 100 from SHHQ dataset~\cite{fu2022stylegan} and 120 collected from the Internet featuring varied poses, outfits, and styles. For in-the-wild images, we first employ the off-the-shelf background removal tool from \href{https://github.com/danielgatis/rembg}{https://github.com/danielgatis/rembg} to obtain human foreground masks, and further create human images with pure white backgrounds, where the subject is centered and occupies approximately 90\% of the image height. All the images are resized to the resolution of 512$\times$512.

\myparagraph{Training details.} 
We train \methodname\ on 8 Nvidia A100 GPUs with 40GB of memory for 1.5 days. The training stage is divided into two stages. In the first stage, the reference UNet and the denoising UNet excluding both 1D and 3D multi-view attention are trained for 30,000 steps with a learning rate of 1e-5 and a total batch size of 64.
A randomly selected novel view image of each subject serves as the target. The dual branches are included, and it takes $\sim$1 day for the first stage. 
In the second stage, we incorporate hybrid 1D-3D attention, training only these parameters for 15,000 steps with a leaning rate of 1e-5 and a total batch size of 8. All 20 views of a subject are used as training targets, and it takes $\sim$0.5 day for the second stage.

\myparagraph{Reconstruction.}
3D human meshes are reconstructed using NeuS~\cite{wang2021neus} following Wonder3D~\cite{long2024wonder3d}, and the vertex colors of the extracted mesh are further optimized with LPIPS loss by differentiable rendering.

\myparagraph{Evaluation of novel view synthesis results.} As shown in Tab. 1 of the main paper, we report both ``4 views'' and ``20 views'' results for novel view synthesis. ``4 views'' are selected as the four orthogonal views with azimuths of $\{0^\circ, 90^\circ, 180^\circ, 270^\circ \}$, which are shared across all methods. To evaluate the quality of dense multi-view generation, we generate ``20 views'' with $18^\circ$ azimuth intervals aligned with our method for all approaches, except Wonder3D~\cite{long2024wonder3d} and SyncDreamer~\cite{liu2023syncdreamer} for they are trained to fixed camera views different from ours.

\section{Discussion}
\subsection{Limitations and Future Works}
\label{sec:supp-limitations}
\myparagraph{Texture details of reconstructed meshes.} 
Although impressive consistency in both geometry and textures has been achieved through our proposed hybrid multi-view attention and geometry-aware dual branch, generating absolutely consistent textures at the pixel level remains challenging. 
This limitation results in reconstructed mesh textures that are not as clear as desired.
This issue is partly due to the denoising process in the latent space, where consistency in low-resolution latents does not fully guarantee absolute consistency in the high-resolution pixel space. Additionally, there is a significant domain gap between 2D images and 3D subjects, and more effective methods for endowing 2D diffusion models with 3D awareness remain to be explored. From the reconstruction perspective, we simply use NeuS~\cite{wang2021neus} for multi-view reconstruction following~\citeauthor{long2024wonder3d}, which requires strict multi-view consistency. Techniques like SDS and image-level losses can reduce the reliance on multi-view consistency for reconstruction quality, which could be further explored to generate meshes with sharper textures based on our proposed human multi-view diffusion model.

\myparagraph{Quality of hands and faces.} 
Due to inherent limitations of our used SD1.5~\cite{rombach2022high} backbone, which struggles to generate high-quality hands and faces in full-body portraits, our generated novel view images often exhibit unrealistic or inconsistent hands and faces as shown in Fig.~\ref{fig:supp-comp-nvs-1}-\ref{fig:supp-comp-nvs-3}. Utilizing more powerful backbones generating at higher resolutions, e.g., SDXL~\cite{podell2023sdxl} and SD3~\cite{esser2024scaling}, can partially alleviate this issue. Additionally, employing specialized techniques for certain parts of human figures could help to produce more detailed results. However, the accurate depiction of human hands and faces remains a significant challenge in the field of image generation.

\subsection{Ethical Consideration.}
\label{sec:supp-ethical}
Human novel view generation and 3D human reconstruction from a single image raise concerns such as privacy violations, intellectual property rights infringement, and potentially improper use for creating ``deep fakes''. Tackling these challenges requires a collective effort to develop ethical guidelines and legal standards. However, we still believe that the proper use of this technique will enhance the research of artificial intelligence and digital entertainment. In this work, considering the sensitivity of personal information, all processed data, models, and results will be strictly used for academic purposes and will not be authorized for commercial use.

\section{More Qualitative Results}
\label{sec:supp-results}
\begin{figure*}[t]
\centering
\includegraphics[width=.95\textwidth]{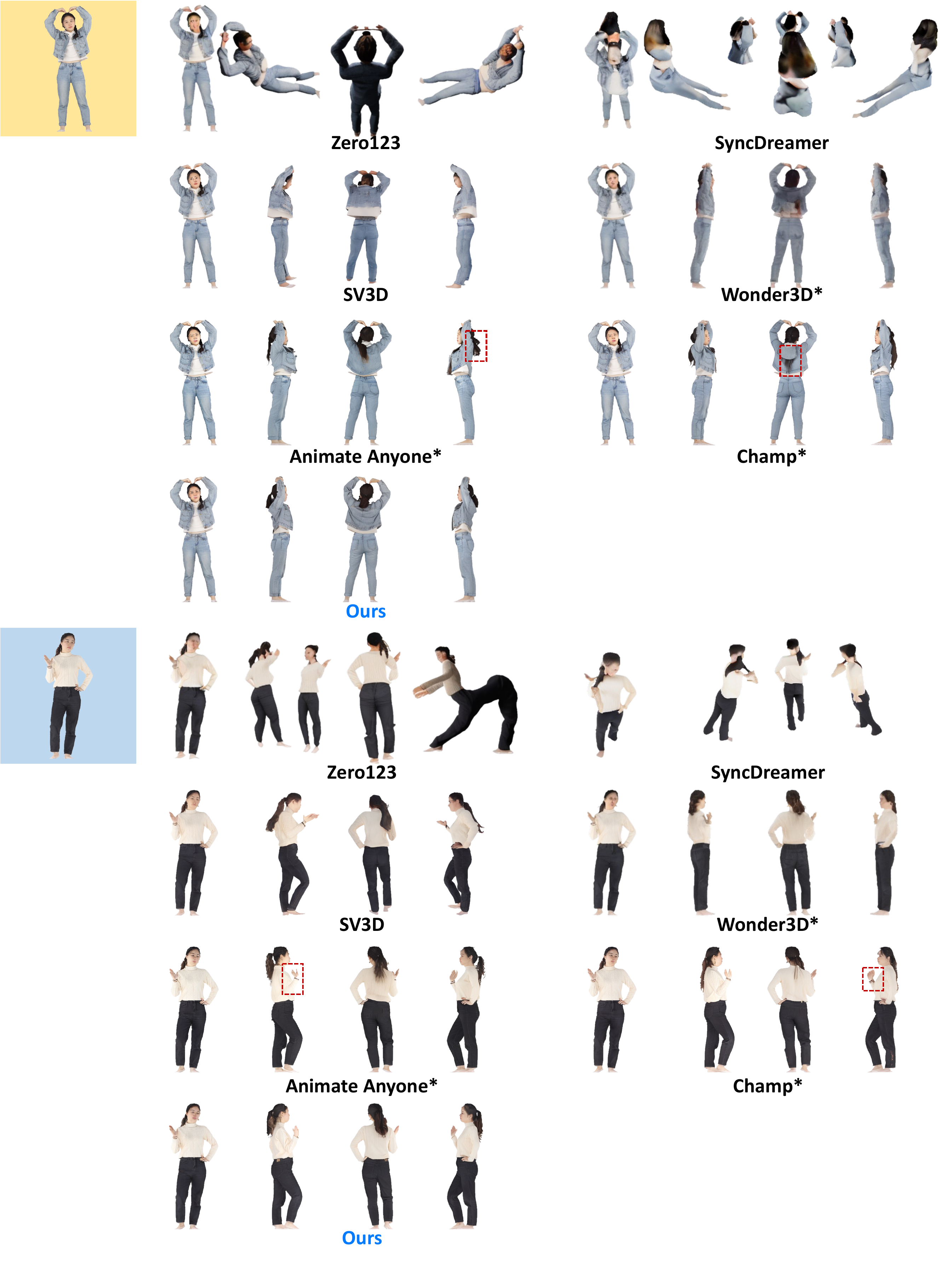}
\caption{Qualitative comparison of novel view synthesis results on THuman2.1~\shortcite{tao2021function4d}. * Methods are finetuned on THuman2.1 dataset for fair comparison.
}
\label{fig:supp-comp-nvs-1}
\end{figure*}

\begin{figure*}[t]
\centering
\includegraphics[width=.95\textwidth]{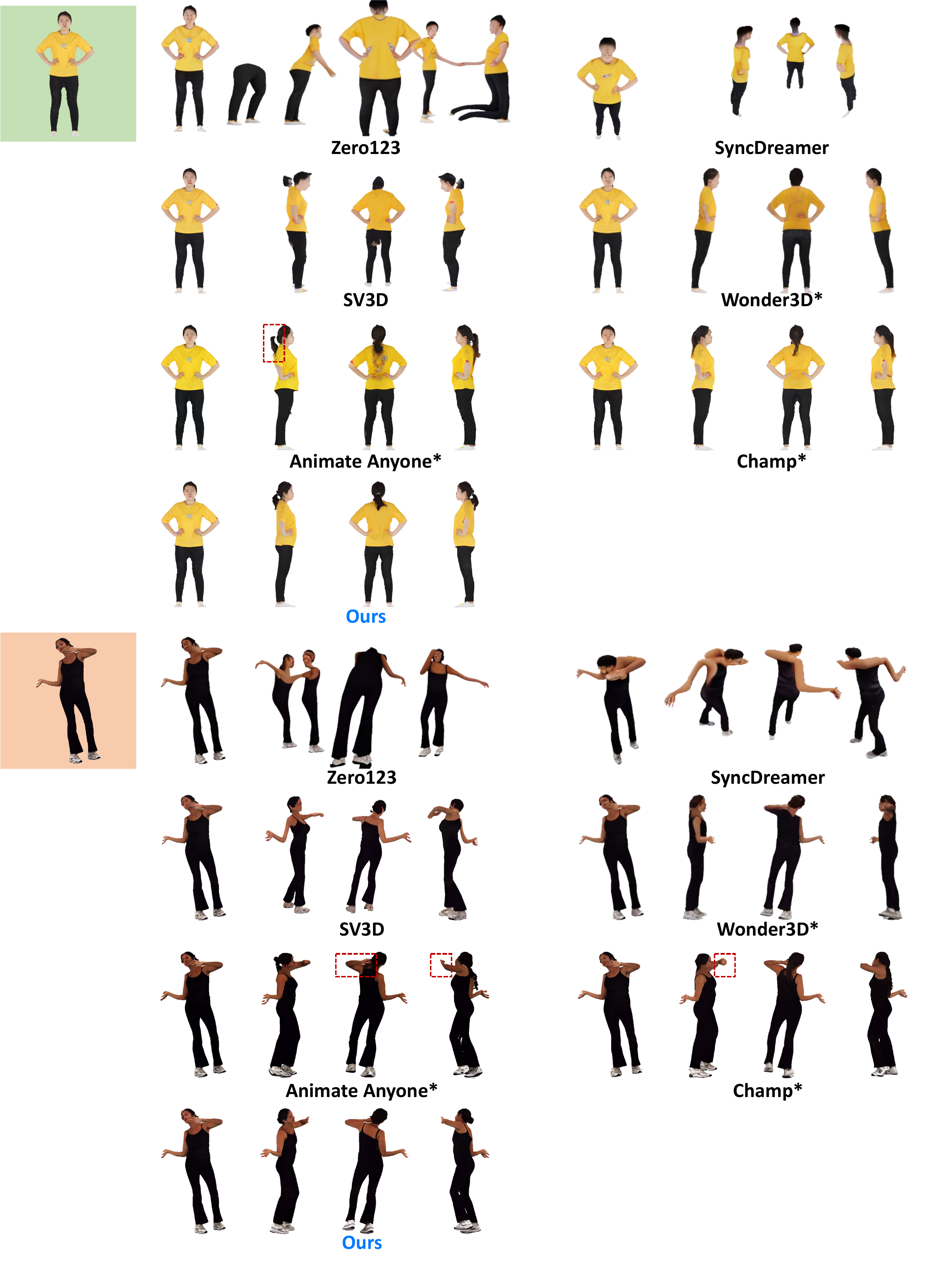}
\caption{Qualitative comparison of novel view synthesis results on THuman2.1~\shortcite{tao2021function4d} (upper) and CustomHumans~\shortcite{ho2023learning} (lower). * Methods are finetuned on THuman2.1 dataset for fair comparison.
}
\label{fig:supp-comp-nvs-2}
\end{figure*}

\begin{figure*}[t]
\centering
\includegraphics[width=.95\textwidth]{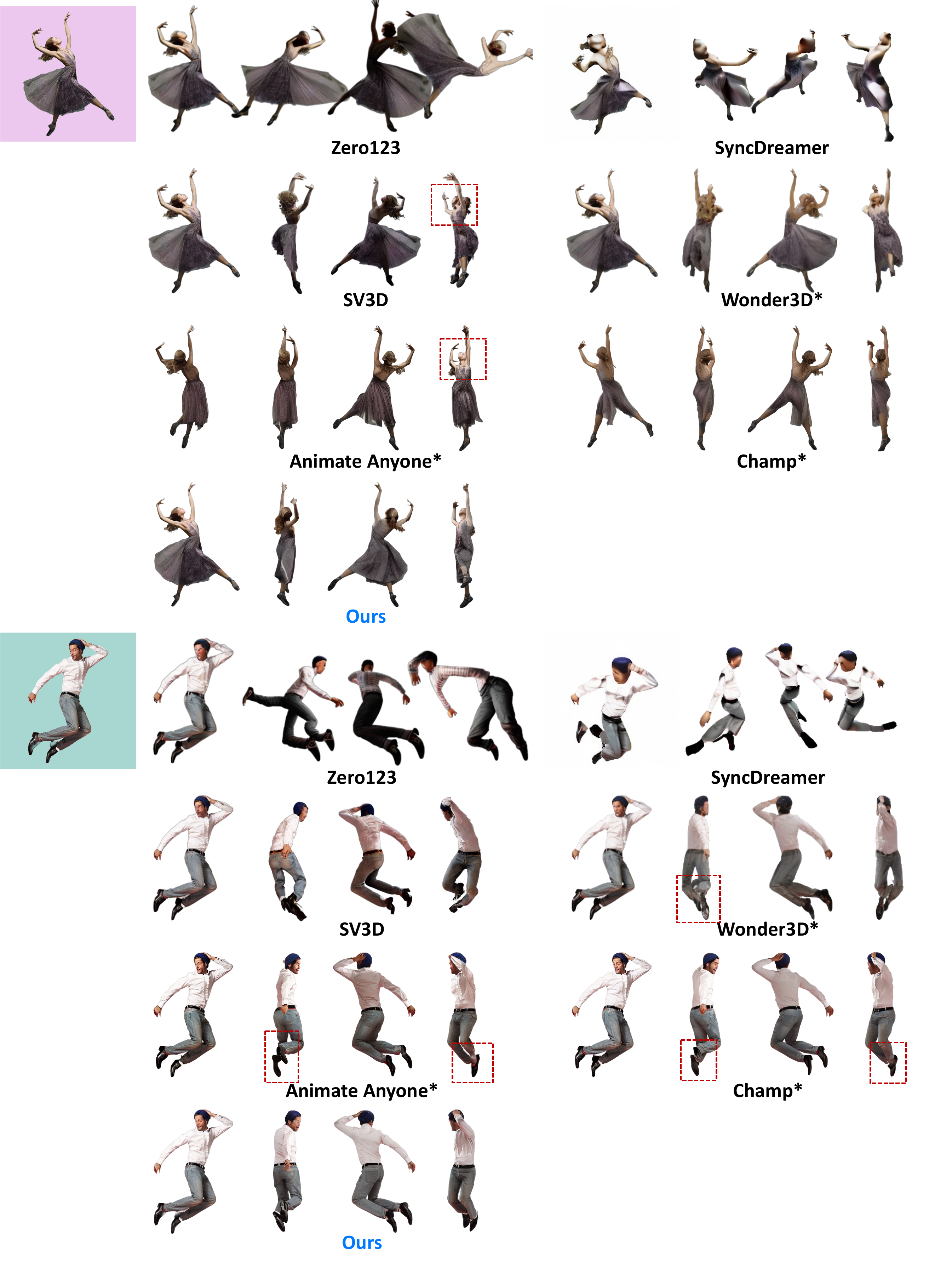}
\caption{Qualitative comparison of novel view synthesis results on in-the-wild data. * Methods are finetuned on THuman2.1 dataset for fair comparison.
}
\label{fig:supp-comp-nvs-3}
\end{figure*}

\myparagraph{Novel view synthesis.} Fig.~\ref{fig:supp-comp-nvs-1}-\ref{fig:supp-comp-nvs-3} present additional novel view results in comparison with state-of-the-art methods, demonstrating that \methodname{} exhibits generalizable performance in human novel view generation with superior quality and enhanced multi-view consistency. 

\begin{figure*}[t]
\centering
\includegraphics[width=.93\textwidth]{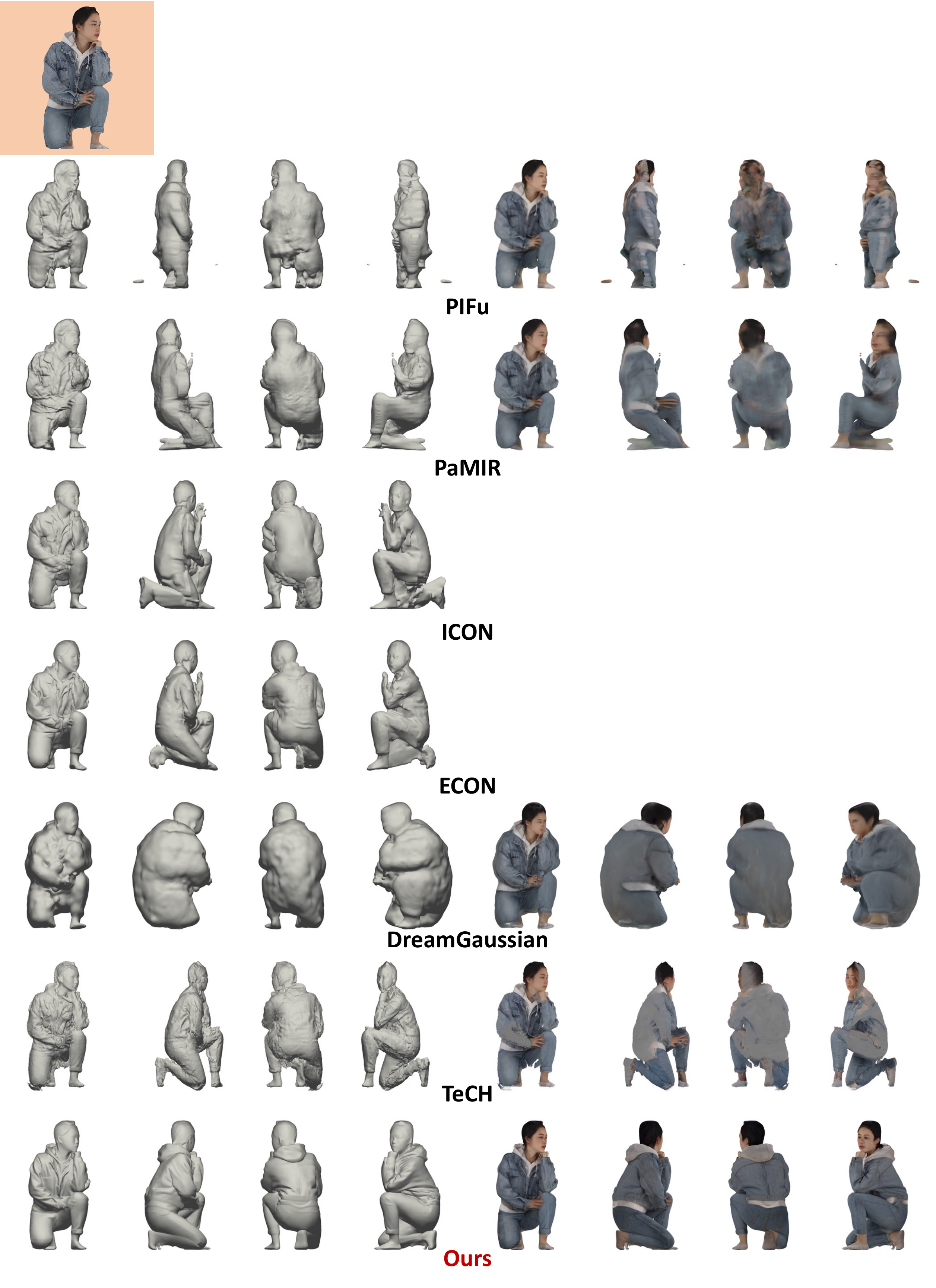}
\caption{Qualitative comparison of 3D human reconstruction results on THuman2.1~\shortcite{tao2021function4d}. 
}
\label{fig:supp-comp-rec-1}
\end{figure*}

\begin{figure*}[t]
\centering
\includegraphics[width=.93\textwidth]{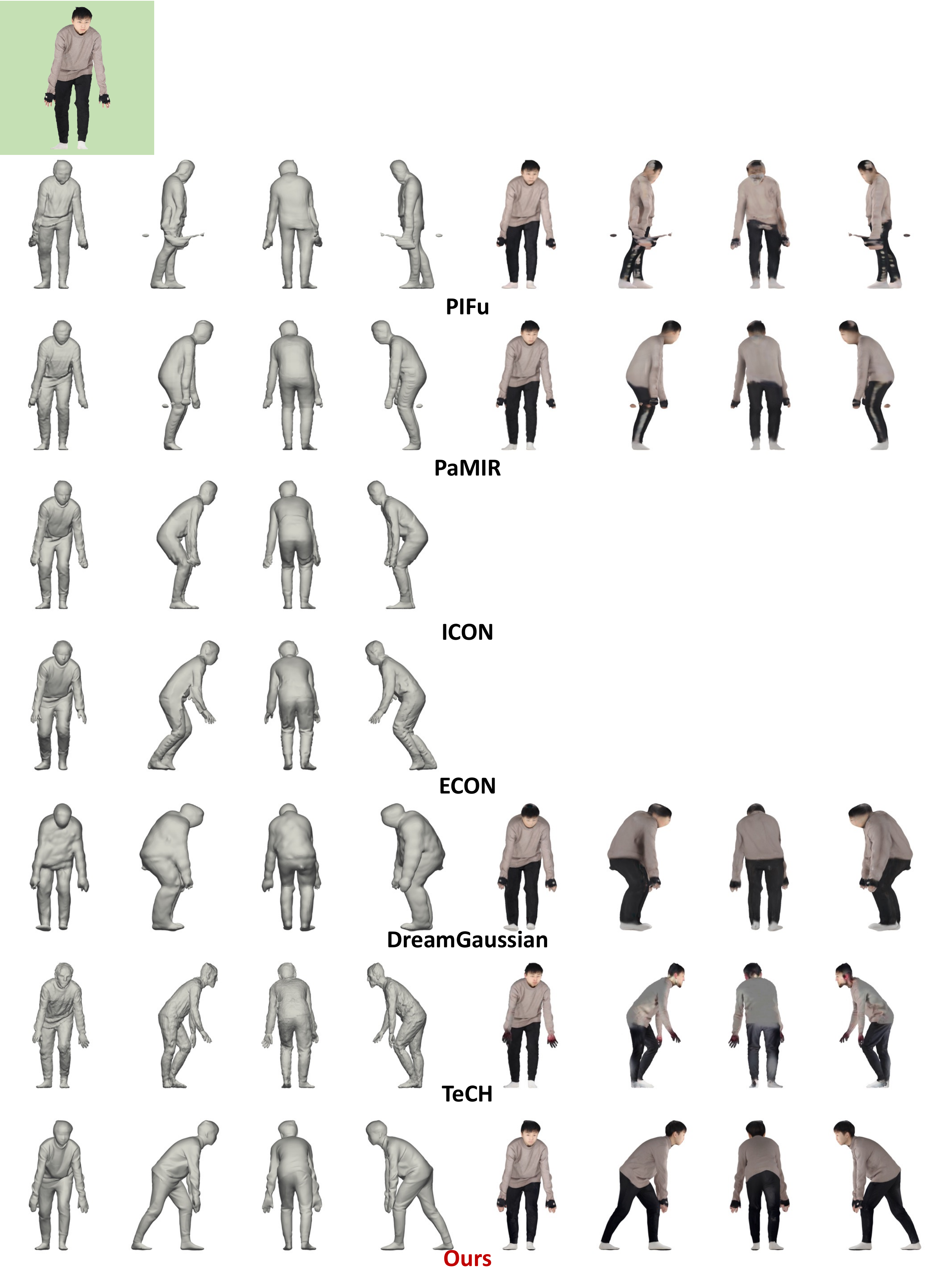}
\caption{Qualitative comparison of 3D human reconstruction results on THuman2.1~\shortcite{tao2021function4d}. 
}
\label{fig:supp-comp-rec-2}
\end{figure*}

\begin{figure*}[t]
\centering
\includegraphics[width=.93\textwidth]{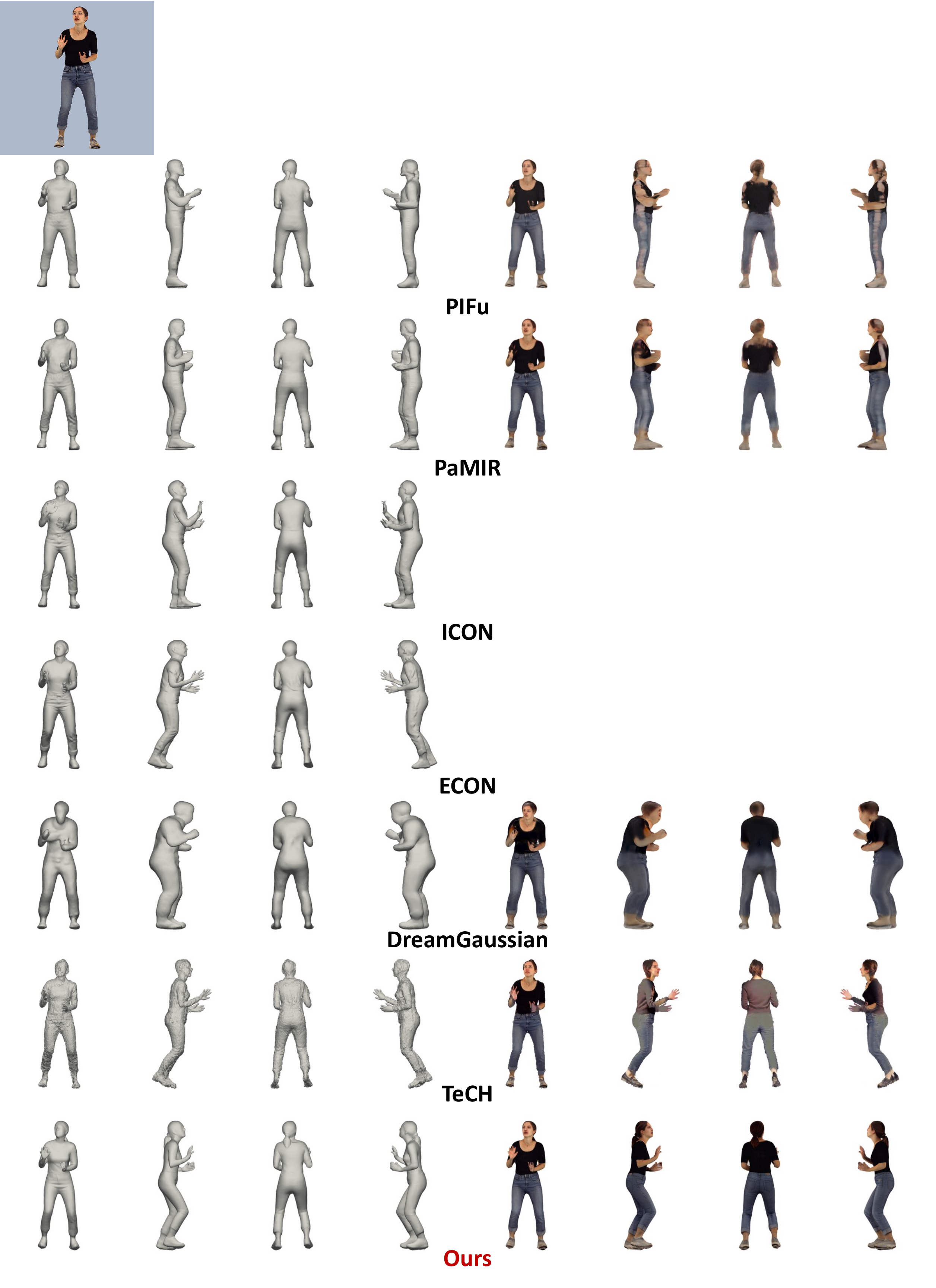}
\caption{Qualitative comparison of 3D human reconstruction results on CustomHumans~\shortcite{ho2023learning}. 
}
\label{fig:supp-comp-rec-3}
\end{figure*}

\begin{figure*}[t]
\centering
\includegraphics[width=.93\textwidth]{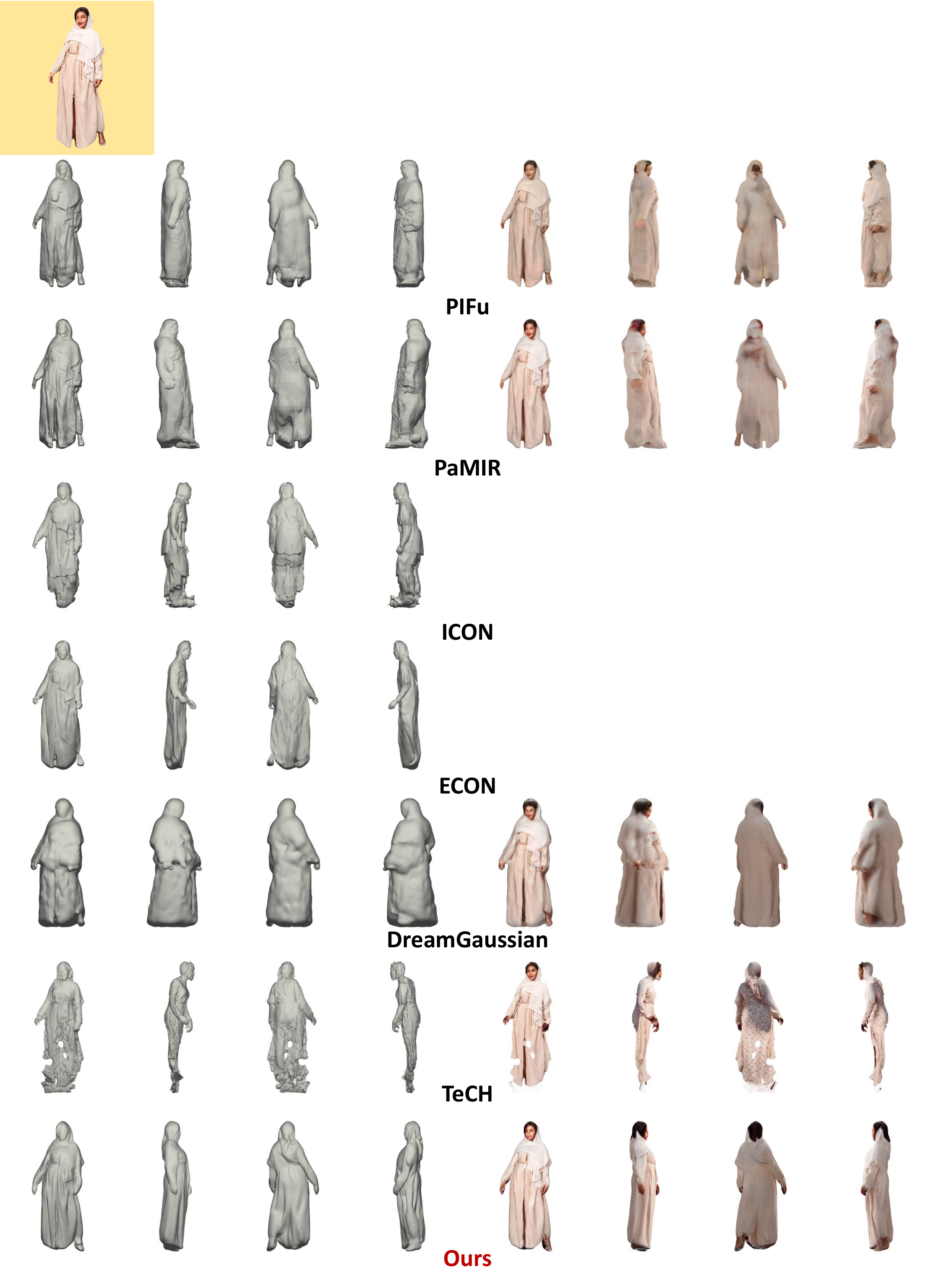}
\caption{Qualitative comparison of 3D human reconstruction results on in-the-wild data. 
}
\label{fig:supp-comp-rec-4}
\end{figure*}

\begin{figure*}[t]
\centering
\includegraphics[width=.93\textwidth]{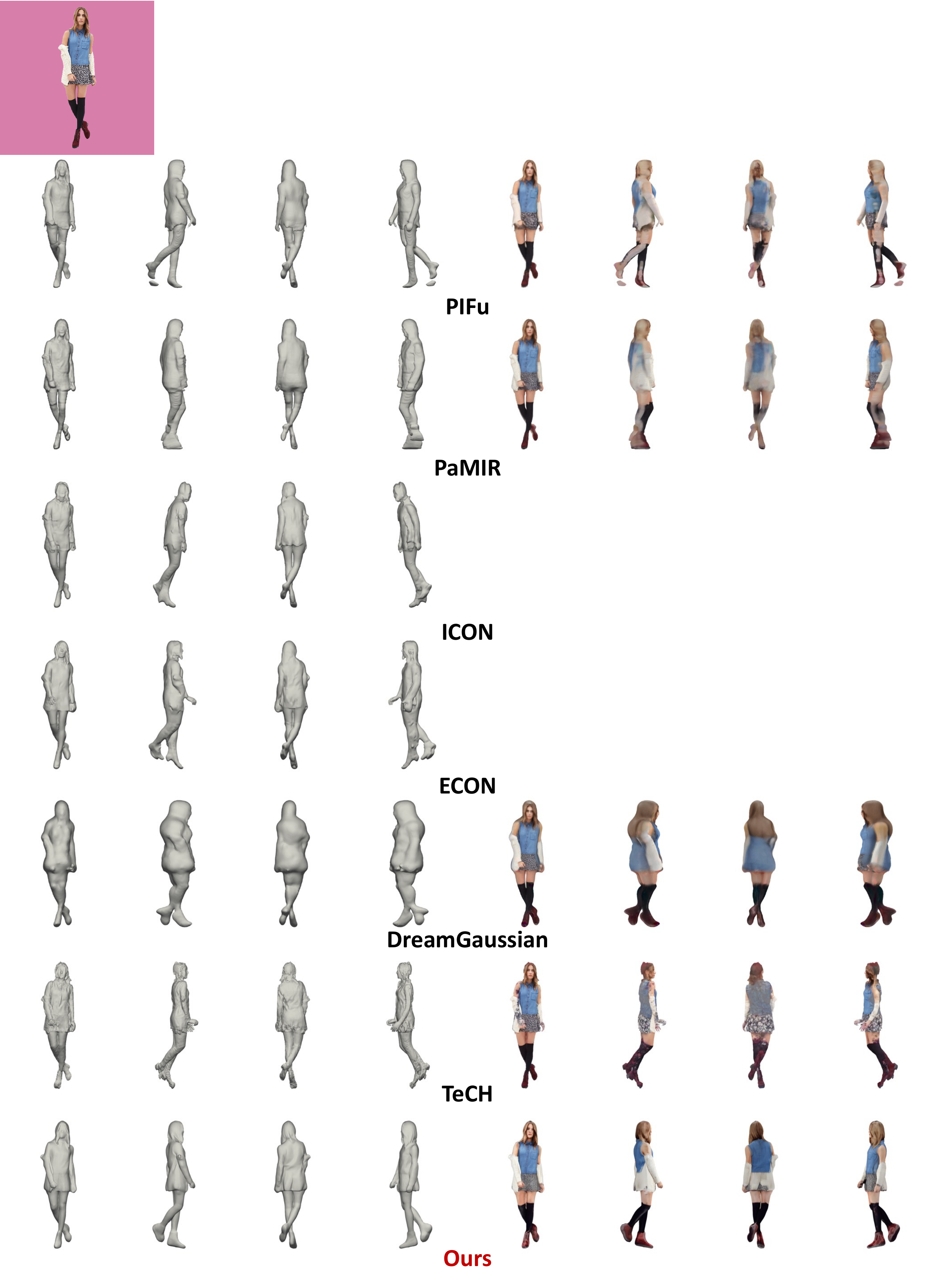}
\caption{Qualitative comparison of 3D human reconstruction results on in-the-wild data. 
}
\label{fig:supp-comp-rec-5}
\end{figure*}

\begin{figure*}[t]
\centering
\includegraphics[width=.93\textwidth]{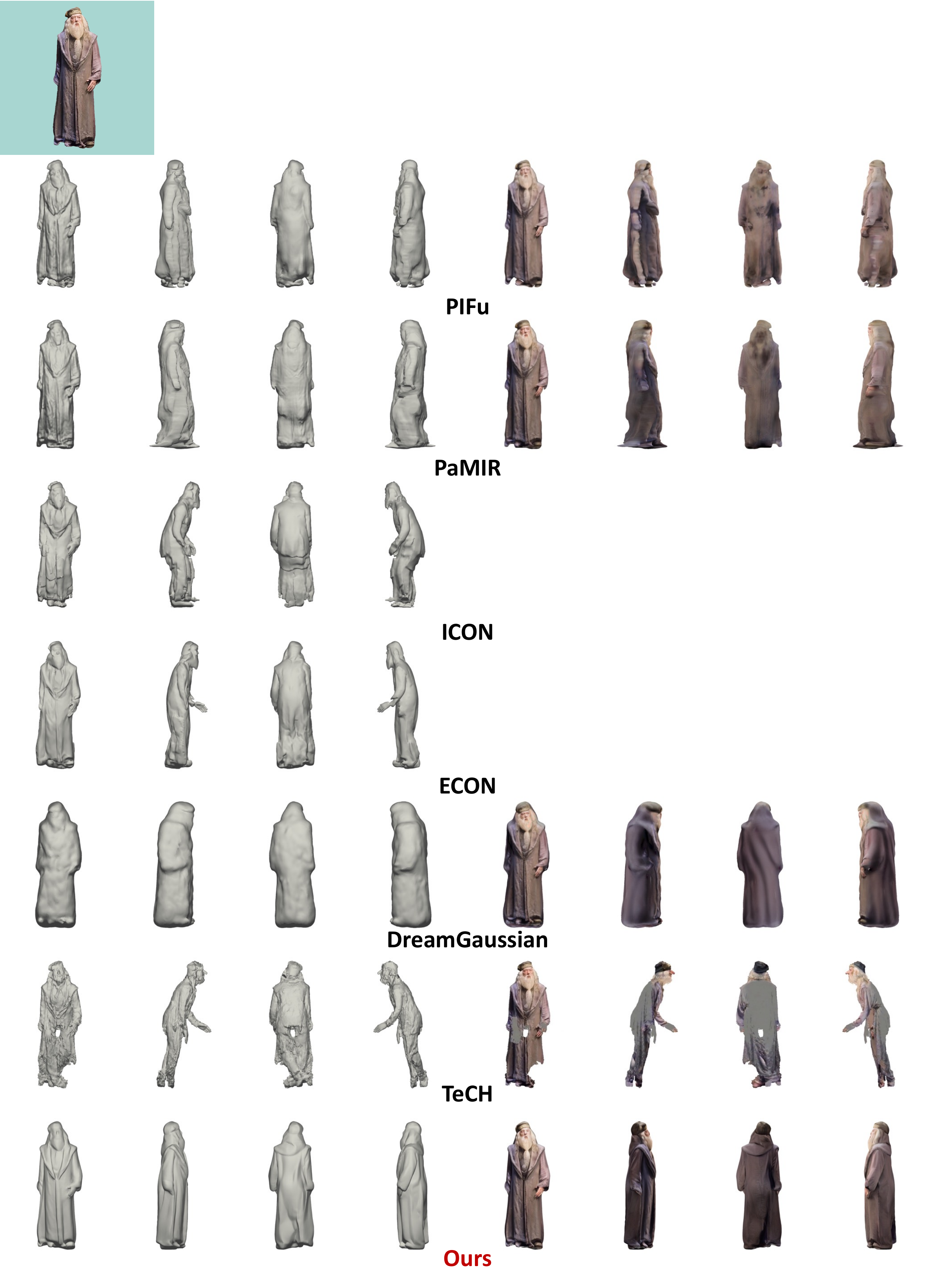}
\caption{Qualitative comparison of 3D human reconstruction results on in-the-wild data. 
}
\label{fig:supp-comp-rec-6}
\end{figure*}

\myparagraph{3D human reconstruction.} Fig.~\ref{fig:supp-comp-rec-1}-\ref{fig:supp-comp-rec-6} showcase additional 3D human reconstruction results and demonstrate that \methodname{} recovers high-quality textured human meshes across a wide range of scenarios, including those with challenging poses and loose clothing.

More qualitative videos are available at \projectpage.

